\documentclass{article}

\PassOptionsToPackage{numbers,sort&compress}{natbib}
\usepackage[preprint]{neurips_2026}

\usepackage[utf8]{inputenc} 
\usepackage[T1]{fontenc}    
\usepackage{hyperref}       
\usepackage{url}            
\usepackage{booktabs}       
\usepackage{amsfonts}       
\usepackage{nicefrac}       
\usepackage{microtype}      
\usepackage{xcolor}         

\usepackage{graphicx}
\usepackage{multirow}

\usepackage{amsmath}
\usepackage{amssymb}
\usepackage{mathtools}
\usepackage{amsthm}

\usepackage{subcaption}
\usepackage{wrapfig}

\usepackage{algorithm}
\usepackage{algorithmic}

\usepackage[table]{xcolor}
\definecolor{lg}{gray}{0.9}  

\newcommand{\ourmethod}{SAVAA}

\definecolor{mydarkred}{rgb}{0.6,0,0}
\definecolor{mydarkgreen}{rgb}{0,0.6,0}
\hypersetup{
	colorlinks=true,
	urlcolor=magenta,
	citecolor=mydarkgreen,
}

\title{SAVAA: Mitigating Hallucinations in LVLMs via Step-wise Adaptive Visual Attention Amplification}

\author{%
Jiacheng Zhang$^{12}$\footnotemark[1],~Feng Liu$^2$,~Chao Du$^1$,~Tianyu Pang$^1$\footnotemark[2] \\
$^1$Sea AI Lab, $^2$The University of Melbourne \\
}

\begin{document}

\maketitle

\footnotetext[1]{This work was done during Jiacheng Zhang’s associate membership at Sea AI Lab.}
\footnotetext[2]{Correspondence to: Tianyu Pang <tianyupang3@gmail.com>}

\begin{abstract}
A line of recent training-free methods for mitigating hallucinations in \emph{large vision-language models} (LVLMs) operates by amplifying attention to visual tokens during autoregressive generation within a single forward pass. We refer to this paradigm as \emph{visual attention amplification} (VAA).
In this paper, we identify a dual failure pattern in existing VAA methods caused by their use of a fixed amplification factor across generation steps: it can be too weak at some steps, leaving hallucinations unresolved, while too strong at others, introducing new hallucinations.
Motivated by this finding, we propose \emph{\textbf{S}tep-wise \textbf{A}daptive \textbf{V}isual \textbf{A}ttention \textbf{A}mplification} (\ourmethod), a new VAA framework that estimates hallucination risk for each generated token and uses the estimated risk to adaptively amplify visual attention at the next generation step.
Specifically, we introduce \emph{\textbf{V}isual \textbf{G}rounding \textbf{E}ntropy} (VGE), a lightweight hallucination-risk estimator that augments predictive entropy with visual grounding, assigning higher risk to tokens that are uncertain, weakly grounded in the image, or both.
Guided by VGE, \ourmethod~uses the estimated risk to calibrate the VAA factor for the next generation step, applying stronger amplification to higher-risk steps and weaker amplification to lower-risk steps.
Across LLaVA-NeXT-7B, Qwen3-VL-8B, and InternVL3.5-8B, \ourmethod~significantly outperforms baseline methods on generative hallucination benchmarks such as CHAIR, SHR and AMBER.
Code is available at: \url{https://github.com/JiachengZ01/SAVVA}.
\end{abstract}

\section{Introduction}
Hallucination is a long-standing challenge in \emph{large vision-language models} (LVLMs) \citep{Huang2023ASO, Liu2023ImprovedBW, Bai2024HallucinationOM}.
LVLM hallucinations arise from multiple factors \citep{Liu2024ASO, Liu2024PayingMA, Zhang2024DebiasingML, Bai2024HallucinationOM, Wu2025LanPRT, Sun2025ExploringCA}. 
One important contributor during autoregressive generation is the model's over-reliance on language priors, especially when visual evidence is under-utilized \citep{Liu2024PayingMA, Zhang2024DebiasingML, Wu2025LanPRT}.
This over-reliance can bias LVLMs toward plausible but visually unsupported content, resulting in outputs inconsistent with the visual input.

To mitigate this issue, a line of recent methods directly amplifies attention to visual tokens during autoregressive generation, a paradigm we refer to as \emph{visual attention amplification} (VAA) \citep{Liu2024PayingMA, Yin2025ClearSightVS, Xie2025TARACMH, Zhao2025TellMW}.
Unlike many hallucination mitigation methods requiring additional training \citep{Zhao2023BeyondHE, yue2024less, xiao2024seeing, xing2024mitigating, sun2024aligning, Zhao2024LookingBT, Peng2025MitigatingOH, Wang2025ImageTM} or multiple forward passes \citep{Huang2023OPERAAH, Leng2023MitigatingOH, Wang2024MitigatingHI, Huo2024SelfIntrospectiveDA, Jiang2023HallucinationAC, Che2025HallucinatoryIT, Fang2025GroundingLW, Wu2025GenerateBV, Chen2025DecouplingCD}, VAA methods operate within a single forward pass without additional training, making them efficient, practical, and readily applicable to existing LVLMs for inference-time deployment \citep{Yin2025ClearSightVS, Xie2025TARACMH, Zhao2025TellMW}.

\begin{figure}[t]
    \centering
    \begin{subfigure}[b]{0.45\linewidth}
        \includegraphics[height=4cm]{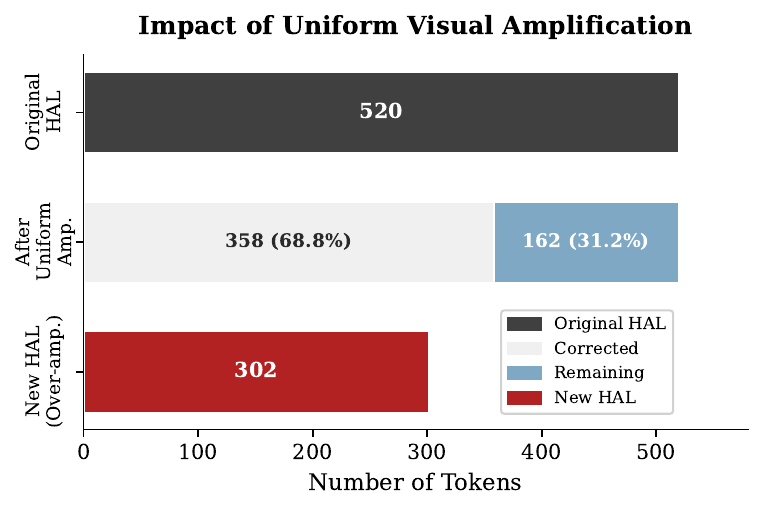}
        \caption{}
        \label{fig: motivation-a}
    \end{subfigure}
    \begin{subfigure}[b]{0.33\linewidth}
        \includegraphics[height=4cm]{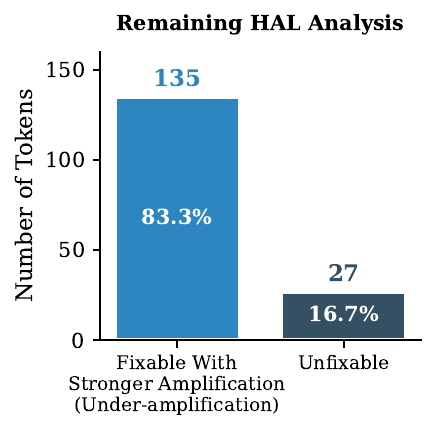}
        \caption{}
        \label{fig: motivation-b}
    \end{subfigure}
    \begin{subfigure}[b]{0.2\linewidth}
        \includegraphics[height=4cm]{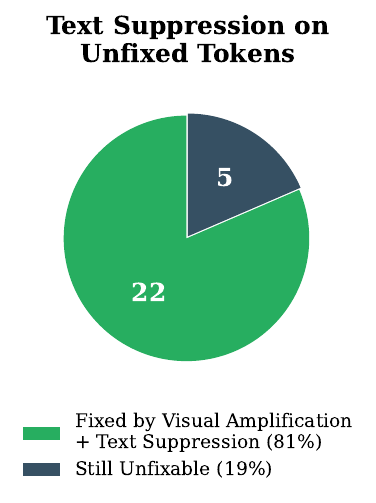}
        \caption{}
        \label{fig: motivation-c}
    \end{subfigure}
    \caption{The proof-of-concept experiment on 200 randomly sampled AMBER examples using LLaVA-NeXT-7B with a fixed VAA factor of 1.2 \citep{Liu2024PayingMA}. GPT-5-mini is used as the token-level hallucination judge. \emph{(a)} Uniform visual amplification corrects many hallucinated tokens, but also leaves unresolved hallucinations and introduces new ones, revealing both under- and over-amplification. \emph{(b)} Most unresolved hallucinations can be fixed by stronger visual amplification, indicating insufficient amplification at the fixed scale. \emph{(c)} For the remaining unfixed tokens, additionally suppressing attention to text tokens resolves most cases. These observations motivate us to design a step-wise VAA factor calibration method with complementary text attention suppression.}
    \label{fig: motivation}
    \vspace{-15pt}
\end{figure}

\begin{wrapfigure}{r}{0.5\linewidth}
\vspace{-15pt}
\centering
\includegraphics[width=\linewidth]{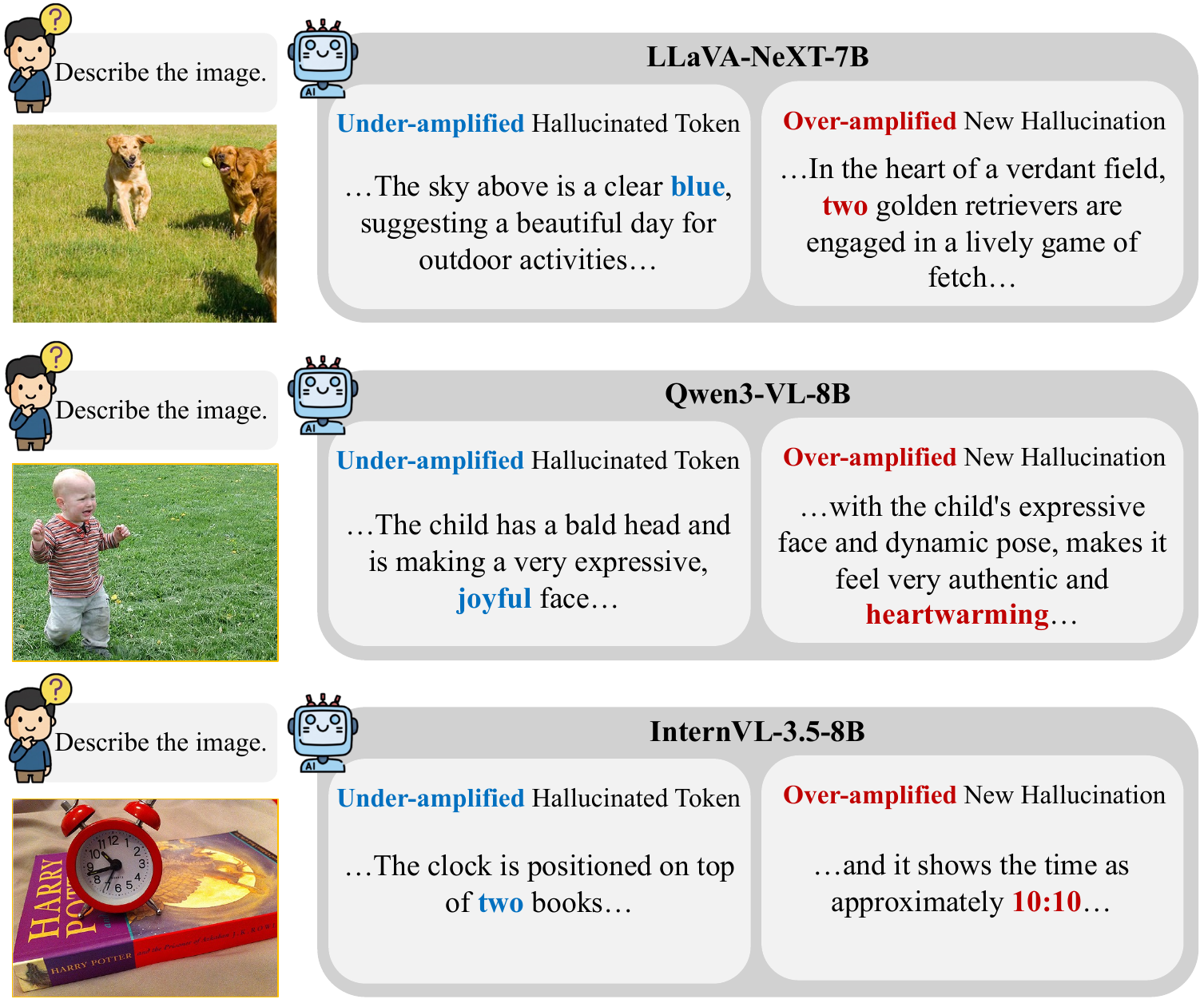}
\caption{Examples of under-amplified and over-amplified hallucinated tokens in LLaVA-NeXT-7B, Qwen3-VL-8B and InternVL3.5-8B.}
\label{fig: motivation example}
\vspace{-15pt}
\end{wrapfigure}

Existing VAA methods mainly differ in which visual tokens they amplify, yet typically keep the amplification factor fixed across generation steps, overlooking how strongly visual attention should be amplified at each step.
To further investigate, we conduct a proof-of-concept experiment using GPT-5-mini \citep{openai2025gpt5mini} as a token-level hallucination judge.
We provide full prompts and judging justifications in Appendix~\ref{A: definition of hallucination}.
As illustrated in Figures~\ref{fig: motivation-a} and~\ref{fig: motivation-b}, this experiment shows that the step-invariant amplification design induces a dual failure pattern:
(1) On the one hand, a fixed amplification factor can be insufficient at some generation steps, leaving a portion of hallucinations unresolved.
(2) On the other hand, the same factor can be excessive at other steps, introducing new hallucinations that are absent from the original response.
These under- and over-amplified cases can even co-exist within the same response and across different LVLM architectures (see Figure~\ref{fig: motivation example}), indicating that the appropriate VAA factor varies across generation steps.
However, heuristically selecting this factor at each generation step is infeasible, motivating us to make a first attempt toward designing an adaptive framework that calibrates the VAA factor step by step.

To this end, we propose \emph{\textbf{S}tep-wise \textbf{A}daptive \textbf{V}isual \textbf{A}ttention \textbf{A}mplification} (\ourmethod), a new VAA framework that estimates hallucination risk for each generated token and uses the estimated risk to adaptively amplify visual attention at the next generation step.
To realize this risk-guided calibration, \ourmethod~requires a lightweight hallucination-risk estimator during autoregressive generation.
A natural choice is predictive entropy, which reflects the model's uncertainty from output logits \citep{Kuhn2023SemanticUL, Kang2025UncertaintyQF}.
However, entropy fails when the model is highly confident but weakly grounded in the image (see Figure~\ref{fig: motivation-vg}).
We therefore introduce \emph{\textbf{V}isual \textbf{G}rounding \textbf{E}ntropy} (VGE), which augments predictive entropy with visual grounding to assign higher risk to generated tokens that are uncertain, weakly grounded in the image, or both (see Section~\ref{sec: vge} and Figure~\ref{fig: motivation-vge}).
Guided by VGE, \ourmethod~uses the estimated risk to calibrate the VAA factor for the next generation step, assigning stronger amplification to higher-risk steps and weaker amplification to lower-risk steps.
The lightweight design of VGE preserves the inference-time efficiency of \ourmethod~(see Section~\ref{sec: compute resource}).

To better understand the remaining failure cases in Figure~\ref{fig: motivation-b}, we investigate whether hallucinations unresolved by visual amplification alone can be addressed by reducing reliance on text input (see Figure~\ref{fig: motivation-c}).
We find that many such tokens become non-hallucinatory when attention to text input tokens is suppressed, suggesting that text attention suppression can reduce residual language-prior dominance beyond visual amplification alone.
Motivated by this finding, \ourmethod~combines step-wise VAA factor calibration with lightweight text attention suppression during generation.
Rather than replacing VAA, this complementary component targets residual hallucinations that remain under stronger visual amplification by weakening language-prior dominance and increasing the relative influence of visual evidence.
We provide ablation studies of text attention suppression in Section~\ref{sec: ablation}.

We comprehensively evaluate \ourmethod~on state-of-the-art hallucination benchmarks, including CHAIR \citep{Rohrbach2018ObjectHI}, SHR \citep{Zhao2023BeyondHE},  AMBER \citep{Wang2023AnLM} and POPE \citep{Li2023EvaluatingOH}, and demonstrate its effectiveness in Section \ref{sec: experiment}. 
Across LLaVA-NeXT-7B \citep{liu2024llavanext}, Qwen3-VL-8B \citep{Bai2025Qwen3VLTR}, and InternVL3.5-8B \citep{Wang2025InternVL35AO}, \ourmethod~consistently outperforms baseline methods by a notable margin, highlighting the importance of moving beyond fixed visual amplification toward step-wise attention calibration during autoregressive generation.

In summary, our contributions are fourfold.
First, we identify a dual failure pattern in existing VAA methods, showing that a fixed amplification factor can be insufficient at some generation steps while excessive at others.
Second, we propose \ourmethod, a new VAA framework that uses VGE to estimate hallucination risk for each generated token and calibrates the VAA factor step-wise during autoregressive generation.
Third, we show that text attention suppression can complement VAA by reducing residual language-prior dominance, and \ourmethod~further incorporates this mechanism during generation.
Fourth, we validate \ourmethod~across three LVLMs on multiple hallucination benchmarks, demonstrating consistent improvements over existing VAA baselines.

\section{Preliminary and Related Work}
\label{sec: related work} 
In this section, we first introduce the background of LVLMs, including their general structure and inference process. Then, we review VAA methods for hallucination mitigation in LVLMs.

\textbf{LVLM structure.}
Given an input image and text prompt, an LVLM encodes the image into visual tokens $\mathbf{X}_i \in \mathbb{R}^{N_i \times d}$ and the prompt into text tokens $\mathbf{X}_p \in \mathbb{R}^{N_p \times d}$. These tokens are concatenated as $\mathbf{X}=[\mathbf{X}_i;\mathbf{X}_p]$ and fed into the LLM backbone for autoregressive generation.
As autoregressive generation proceeds, the growing number of generated text tokens, in contrast to the fixed set of visual tokens, increasingly dominates the attention distribution, causing the model to place less emphasis on visual information and become more influenced by language priors  \citep{Liu2024PayingMA, Liu2025MoreTL}.

\textbf{LVLM inference.} During inference, LVLMs generate a response sequence $R = \{y_1, \ldots, y_L\}$ in an autoregressive manner.
At each generation step $t$, the model predicts the next token conditioned on the input visual tokens $\mathbf{X}_i$, the input text tokens $\mathbf{X}_p$, and the previously generated tokens $\{y_1, \ldots, y_{t-1}\}$.
At each step, the influence of visual and textual information on next-token prediction is governed by the self-attention mechanism within the LLM.
Specifically, within the $l$-th Transformer layer, causal self-attention is applied over the combined token sequence.
For the newly generated token at step $t$, the query $\mathbf{Q}_l^{(t)}$ attends to all cached keys $\mathbf{K}_l$ via:
\begin{equation}
\mathbf{A}_l^{(t)} = \mathrm{Softmax}\!\left(
\frac{\mathbf{Q}_l^{(t)} \mathbf{K}_l^\top}{\sqrt{d}}
\right), \nonumber
\end{equation}
where $\mathbf{A}_l^{(t)}$ determines how the current token attends to visual tokens, text tokens, and previously generated tokens.
Under this scheme, the attention weights assigned to visual tokens directly control the extent to which visual information influences generation at step $t$.

\textbf{LVLM hallucination mitigation via VAA.} 
Hallucination mitigation methods can be broadly grouped into training-based methods \citep{Zhao2023BeyondHE, yue2024less, xiao2024seeing, xing2024mitigating, sun2024aligning, Zhao2024LookingBT, Peng2025MitigatingOH, Wang2025ImageTM}, decoding-based methods \citep{Huang2023OPERAAH, Leng2023MitigatingOH, Wang2024MitigatingHI, Huo2024SelfIntrospectiveDA, Jiang2023HallucinationAC, Che2025HallucinatoryIT, WangSHIFTSH, Fang2025GroundingLW, Wu2025GenerateBV, Chen2025DecouplingCD}, and \emph{visual attention amplification} (VAA) methods \citep{Liu2024PayingMA, Yin2025ClearSightVS, Xie2025TARACMH, Zhao2025TellMW}.
In this paper, we focus on VAA, a training-free inference-time paradigm that amplifies attention to visual tokens during autoregressive generation within a single forward pass.
Early efforts such as \emph{pay attention to image} (PAI) \citep{Liu2024PayingMA} uniformly amplify attention to all visual tokens.
Building on this idea, subsequent works have largely focused on which visual tokens to amplify by identifying visually relevant tokens or regions during generation.
For example, \emph{visual amplification fusion} (VAF) \citep{Yin2025ClearSightVS} fuses visual signals into the attention process to enhance visually grounded representations.
\emph{Temporal attention real-time accumulation connection} (TARAC) \citep{Xie2025TARACMH} accumulates cross-modal attention patterns over time to identify modality-specific regions that warrant stronger visual emphasis.
Similarly, \emph{vision-guided attention} (VGA) \citep{Zhao2025TellMW} uses grounding-aware signals to identify visually relevant tokens and guide attention toward them.
Despite differing in which visual tokens they amplify, existing VAA methods share a common step-invariant design that uses a predefined amplification factor.
Such a design cannot account for the fact that different generated tokens may require different levels of visual amplification.
To the best of our knowledge, \ourmethod~is the first framework to explicitly formulate risk-guided, step-wise factor calibration during autoregressive generation.

\textbf{Hallucination risk estimation.}
Risk-guided VAA factor calibration requires estimating hallucination risk during autoregressive generation.
Existing hallucination risk estimation methods can be broadly grouped into token-probability-based methods \citep{Takayama2019RelevantAI, Malinin2021UncertaintyEI}, output-consistency-based methods \citep{Zhao2023KnowingWL, Nikitin2024KernelLE}, and internal-state-based methods \citep{Zhang2023EnhancingUH, Sriramanan2024LLMCheckID, Du2024HaloScopeHU}.
Output-consistency-based methods typically require multiple forward passes, while internal-state-based methods require accessing and storing intermediate representations, making them less suitable for lightweight inference-time VAA methods.
In contrast, token-probability-based methods can be computed directly from output logits with negligible overhead.
Within this category, predictive entropy is widely used as a lightweight uncertainty signal for hallucination risk estimation \citep{Malinin2021UncertaintyEI, Kuhn2023SemanticUL, Kang2025UncertaintyQF}.
However, entropy fails when the model is highly confident but weakly grounded in the image.
This limitation motivates VGE, which augments entropy with visual grounding to better estimate hallucination risk while preserving the lightweight nature required for step-wise VAA factor calibration.
We formally introduce VGE in Section~\ref{sec: vge}.

\section{Step-wise Adaptive Visual Attention Amplification}
\label{sec: method}

We propose \ourmethod, a new VAA framework with two key designs: a lightweight VGE risk estimator and a step-wise adaptive VAA factor calibration mechanism.
At each generation step, \ourmethod~estimates hallucination risk after generating the current token and uses this risk to calibrate the VAA factor for the next step.
This section presents \ourmethod~through four components: VGE for hallucination-risk estimation, step-wise VAA factor calibration, and text attention suppression as a complementary mechanism to VAA, followed by the overall attention modulation.
We visually illustrate SAVAA in Figure \ref{fig: pipeline} and provide the full algorithm in Appendix~\ref{A: algo}.

\subsection{Visual Grounding Entropy}
\label{sec: vge}
\begin{figure}[t]
    \centering
    \begin{minipage}[t]{0.55\linewidth}
        \centering
        \includegraphics[height=2.6cm,keepaspectratio]{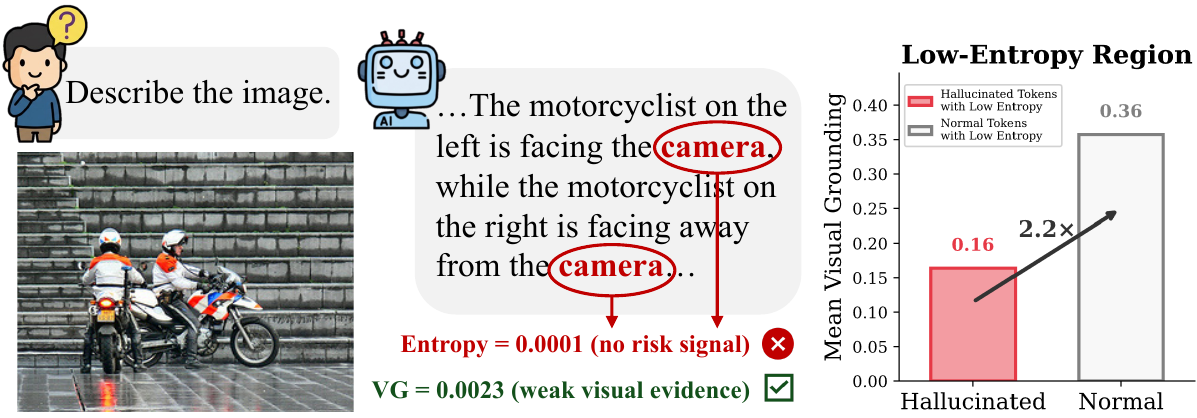}
        \caption{
        Motivation for visual grounding. 
        \emph{Left}: A LLaVA-NeXT-7B example where entropy misses a confident hallucination.
        \emph{Right}: In the low-entropy region, hallucinated tokens have lower visual grounding scores than normal tokens, suggesting that visual grounding provides a complementary risk signal.
        }
        \label{fig: motivation-vg}
    \end{minipage}
    \hfill
    \begin{minipage}[t]{0.43\linewidth}
        \centering
        \includegraphics[height=2.6cm,keepaspectratio]{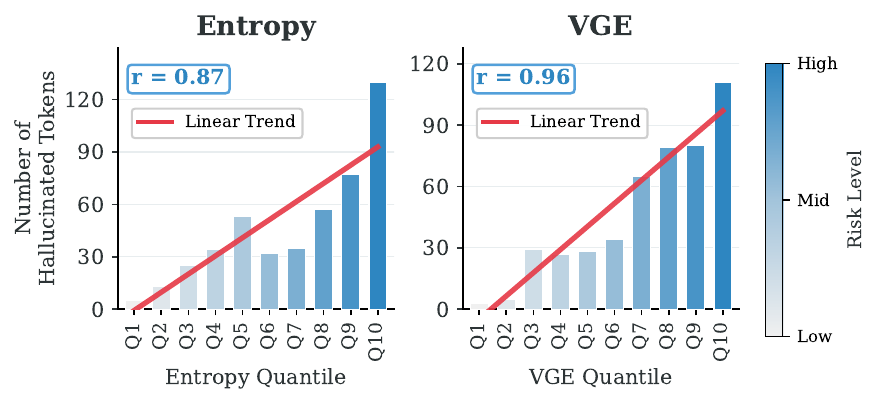}
        \caption{
        Comparison of entropy and VGE as hallucination risk estimators on LLaVA-NeXT-7B.
        VGE shows a stronger correlation with hallucination counts than entropy, indicating improved risk estimation beyond uncertainty alone.
        }
        \label{fig: motivation-vge}
    \end{minipage}
    \vspace{-10pt}
\end{figure}

\textbf{Entropy and its limitation.}
Risk-guided VAA factor calibration requires a lightweight risk signal at each generation step.
A natural choice is predictive entropy, which can be computed directly from the output logits.
Formally, let $\mathcal{V}$ denote the vocabulary with size $V=|\mathcal{V}|$.
At generation step $t$, given the output logits $\mathbf{z}_t \in \mathbb{R}^{V}$, the predictive distribution is
\begin{equation}
p_t(v) = \mathrm{Softmax}(\mathbf{z}_t)_v, \quad v \in \mathcal{V}. \nonumber
\end{equation}
We use the normalized predictive entropy:
\begin{equation}
\label{eq: normalized_entropy}
\bar{H}_t
=
\frac{-\sum_{v \in \mathcal{V}} p_t(v)\log p_t(v)}{\log V}.
\end{equation}
However, entropy alone can underestimate hallucination risk when the model is highly confident but weakly grounded in the image.
For example, Figure~\ref{fig: motivation-vg} shows a LLaVA-NeXT case where the hallucinated token ``camera'' has extremely low entropy, despite lacking visual evidence in the image.
This motivates incorporating visual grounding as a complementary signal.

\textbf{Visual grounding score.}
To provide visual-evidence awareness beyond entropy, we compute a vocabulary-level grounding vector $\mathbf{G}$ once during the prefilling stage.
Given the image $\mathbf{I}$ and text prompt $\mathbf{x}$, the LVLM first encodes the image into visual tokens $\mathbf{X}_i$ and performs prefilling to obtain the hidden states $\mathbf{H}_0$.
We then use the hidden states at visual-token positions to construct the grounding vector.
Let $\mathcal{I}_{\mathrm{visual}}$ denote the set of visual-token indices, and let $\mathbf{h}_i \in \mathbb{R}^{V}$ denote the vocabulary logits obtained by applying the LM head to the prefilling hidden state $\mathbf{H}_0^{(i)}$ of visual token $i \in \mathcal{I}_{\mathrm{visual}}$:
\begin{equation}
\mathbf{h}_i = \mathrm{LMHead}(\mathbf{H}_0^{(i)}), \quad \forall i \in \mathcal{I}_{\mathrm{visual}}. \nonumber
\end{equation}
We define the grounding vector $\mathbf{G} \in \mathbb{R}^{V}$ as:
\begin{equation}
\label{eq: grounding_vector}
\mathbf{G}[v]
=
\operatorname{Pool}_{i \in \mathcal{I}_{\mathrm{visual}}}
\mathrm{Softmax}(\mathbf{h}_i)[v],
\quad \forall v \in \mathcal{V}.
\end{equation}
By default, we use max pooling, i.e., $\operatorname{Pool}=\max$, and evaluate mean and top-$k$ average pooling in the ablation study (see Section \ref{sec: ablation}).
At generation step $t$, let
$v_t^* = \arg\max_{v \in \mathcal{V}} \mathbf{z}_t[v]$
denote the token predicted from the output logits $\mathbf{z}_t$.
We then index the prefilling-stage grounding vector $\mathbf{G}$ by $v_t^*$ to obtain the token-specific visual grounding score:
\begin{equation}
\label{eq: step-wise-G}
G_t = \mathbf{G}[v_t^*].
\end{equation}
A lower $G_t$ indicates weaker visual support for the current prediction.
As shown in Figure~\ref{fig: motivation-vg}, hallucinated tokens in the low-entropy region exhibit lower visual grounding scores than normal tokens, suggesting that visual grounding provides useful evidence-awareness when entropy is uninformative.
Since $\mathbf{G}$ is computed once during prefilling and reused across all decoding steps, this grounding signal introduces no additional forward passes.

\textbf{Combining entropy and visual grounding.}
While the grounding score provides visual-evidence awareness, it is computed from the prefilling-stage grounding vector and remains static throughout decoding.
Thus, grounding alone cannot reflect the model's step-specific uncertainty during autoregressive generation.
We therefore combine visual grounding with predictive entropy and define \emph{\textbf{V}isual \textbf{G}rounding \textbf{E}ntropy} (VGE) as a lightweight hallucination-risk estimator:
\begin{equation}
\label{eq: vge}
\mathrm{VGE}_t
=
\alpha \bar{H}_t
+
(1-\alpha)(1-G_t),
\end{equation}
where $\alpha \in [0,1]$ balances predictive uncertainty and visual grounding.
A larger $\mathrm{VGE}_t$ indicates that the current prediction is uncertain, weakly grounded in the image, or both.

As shown in Figure~\ref{fig: motivation-vge}, VGE provides a more reliable risk-ranking signal than entropy alone.
Entropy yields a less stable ordering of hallucination risk across quantiles, whereas VGE shows a more monotonic increase in hallucination counts from low- to high-risk quantiles.
This suggests that incorporating visual grounding improves risk estimation beyond uncertainty alone, making VGE better suited for risk-guided, step-wise VAA factor calibration.

\textbf{Positioning of VGE.}
Although VGE correlates with hallucinated tokens, we do not position it as a standalone hallucination detector.
Unlike prior detection methods that are typically used for post-hoc analysis \citep{Zhao2023KnowingWL, Zhang2023EnhancingUH, Nikitin2024KernelLE, Sriramanan2024LLMCheckID, Du2024HaloScopeHU}, VGE serves as an online risk signal for one-step-lagged VAA factor calibration during autoregressive generation.
It is computed without additional training or extra forward passes, and the resulting risk is used to set the VAA factor for the next generation step.
This also distinguishes our use of visual grounding from VGA~\citep{Zhao2025TellMW}: VGA uses grounding to decide which visual tokens or regions to amplify, whereas VGE uses grounding to estimate \emph{how strongly} visual attention should be amplified in the subsequent step.
In this sense, VGE gives visual grounding a different functional role from prior VAA methods: rather than selecting visual tokens, it supports risk-guided calibration of the VAA factor.

\begin{figure}[t]
\begin{center}
\includegraphics[width=\textwidth]{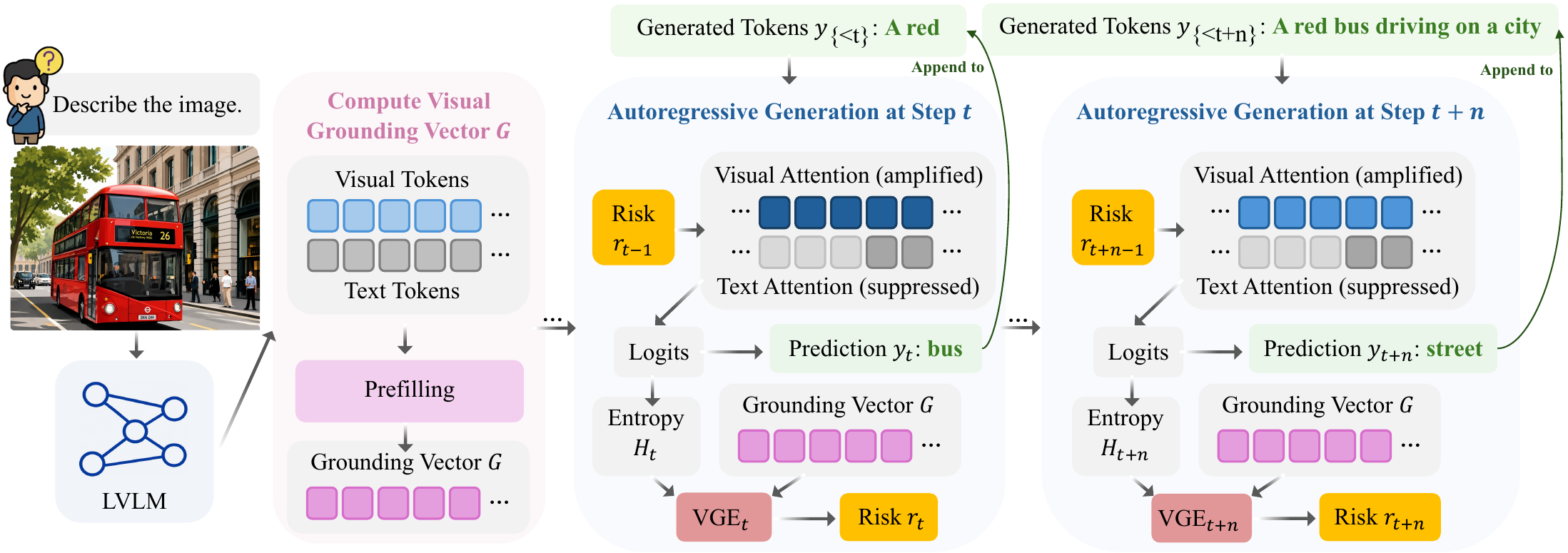} 
\end{center}
\caption{
Overview of \ourmethod.
The LVLM first computes a grounding vector $\mathbf{G}$ during prefilling and reuses it across generation steps.
At generation step \(t\), the previous-step risk \(r_{t-1}\) calibrates the current VAA factor and modulates attention before predicting \(y_t\).
The resulting logits, together with the same grounding vector $\mathbf{G}$, are used to compute the new risk \(r_t\), which is passed to the next step.
}
\label{fig: pipeline}
\vspace{-10pt}
\end{figure}

\subsection{Step-wise Adaptive VAA Factor Calibration}
\label{sec: vaa_calibration}
\textbf{From VGE to risk score.}
After generating token $y_t$, we convert $\mathrm{VGE}_t$ into a normalized hallucination risk score:
\begin{equation}
\label{eq: risk_score}
r_t = \min\left(\frac{\mathrm{VGE}_t}{\gamma}, 1\right),
\end{equation}
where $\gamma$ is a risk scale that controls the sensitivity of risk normalization.
The resulting $r_t \in [0,1]$ reflects the estimated hallucination risk after step $t$, with larger values indicating a stronger need for visual amplification in the subsequent generation step.

\textbf{One-step-lagged VAA factor calibration.}
Since $r_t$ is computed from the logits produced at step $t$, it cannot be used to modulate the attention of the same step.
Therefore, \ourmethod~uses the previous-step risk $r_{t-1}$ to calibrate the VAA factor for the current generation step.
Specifically, at step $t$, we define:
\begin{equation}
\label{eq: vaa_factor}
m_t = 1 + (m^{\max}_{\text{vis}} - 1)\cdot r_{t-1},
\end{equation}
where $m^{\max}_{\text{vis}}$ denotes the maximum VAA factor.
For the first generation step, we initialize $r_0=0$ because no previous generated token is available for risk estimation. Thus, $m_1=1$ and the model starts from its original attention behavior.
When $r_{t-1}=0$, no visual amplification is applied (i.e., $m_t=1$), avoiding unnecessary intervention.
As $r_{t-1}$ increases, $m_t$ grows smoothly, enabling stronger visual amplification after higher-risk predictions.

\textbf{Pre-softmax visual attention amplification.}
We apply the calibrated VAA factor $m_t$ to the pre-softmax self-attention scores associated with visual tokens.
At generation step $t$, within the $\ell$-th Transformer layer, let
$\mathbf{Z}_\ell^{(t)} \in \mathbb{R}^{1 \times (N_i + N_p + t - 1)}$
denote the pre-softmax attention scores for the current token.
The corresponding attention weights are computed as $\mathrm{Softmax}(\mathbf{Z}_\ell^{(t)})$.
For layers $\ell \in [L_s, L_e)$, we modulate the scores assigned to visual tokens as:
\begin{equation}
\label{eq: visual_amplification}
\tilde{\mathbf{Z}}_\ell^{(t)}[-1, \mathcal{I}_{\mathrm{visual}}]
=
m_t \cdot \mathbf{Z}_\ell^{(t)}[-1, \mathcal{I}_{\mathrm{visual}}],
\end{equation}
where $\mathcal{I}_{\mathrm{visual}}$ denotes the index set of visual tokens.

\subsection{Text Attention Suppression}
\textbf{Motivation.}
While step-wise VAA factor calibration strengthens visual evidence, Figure~\ref{fig: motivation-c} shows that some hallucinations remain unresolved even under stronger visual amplification.
These cases suggest that hallucinations can also persist due to residual dominance of the text context or language priors.
Therefore, in addition to amplifying visual-token attention, we introduce a lightweight text attention suppression mechanism as a complementary intervention.

\textbf{Suppressing text attention.}
At generation step $t$, let $\mathcal{I}_{\mathrm{text}}$ denote the index set of input text tokens.
For layers $\ell \in [L_s, L_e)$, we suppress the pre-softmax self-attention scores assigned to input text tokens:
\begin{equation}
\label{eq: text_suppression}
\tilde{\mathbf{Z}}_\ell^{(t)}[-1, \mathcal{I}_{\mathrm{text}}]
=
\frac{1}{m^{\max}_{\text{txt}}}
\cdot
\mathbf{Z}_\ell^{(t)}[-1, \mathcal{I}_{\mathrm{text}}],
\end{equation}
where $m^{\max}_{\text{txt}}$ denotes the maximum text suppression factor.

\textbf{Relation to VAA.}
Text attention suppression is not intended to replace visual amplification.
Instead, it complements VAA by reducing the relative influence of text tokens in the softmax attention distribution.
After softmax normalization, decreasing text attention scores increases the relative contribution of visual evidence, helping mitigate residual language-prior dominance.
Unlike the VAA factor $m_t$, which is calibrated step-wise by hallucination risk, we keep $m^{\max}_{\text{txt}}$ fixed as a lightweight complementary mechanism.
We evaluate its effect in the ablation study in Section~\ref{sec: ablation}.

\subsection{Overall Attention Modulation}
\textbf{Final attention weights}.
After applying step-wise adaptive VAA factor calibration and text attention suppression, the final attention weights are obtained by applying softmax to the modulated attention scores:
\begin{equation}
\tilde{\mathbf{A}}_\ell^{(t)}
=
\mathrm{Softmax}\!\left(\tilde{\mathbf{Z}}_\ell^{(t)}\right). \nonumber
\end{equation}
Through softmax normalization, amplifying visual-token scores increases the attention mass allocated to visual evidence, while suppressing text scores reduces the relative influence of language priors.
Thus, \ourmethod~reallocates attention toward visual evidence through two complementary mechanisms: a risk-guided, one-step-lagged VAA factor \(m_t\) that controls visual amplification at step \(t\), and a lightweight fixed text suppression factor \(m^{\max}_{\mathrm{txt}}\).
This design enables stronger visual reliance after higher-risk predictions while avoiding unnecessary visual amplification after lower-risk ones.

\section{Experiment}
\label{sec: experiment}
\subsection{Experiment Setting}
\label{sec: experiment settings}

\textbf{LVLM architectures.}
Following prior work~\citep{Xie2025TARACMH}, we evaluate \ourmethod~on three LVLMs with different architectural designs: LLaVA-NeXT-7B~\citep{liu2024llavanext}, Qwen3-VL-8B~\citep{Bai2025Qwen3VLTR}, and InternVL3.5-8B~\citep{Wang2025InternVL35AO}.

\textbf{Hallucination benchmarks.}
We evaluate \ourmethod~on four state-of-the-art hallucination benchmarks: CHAIR~\citep{Rohrbach2018ObjectHI}, SHR~\citep{Zhao2023BeyondHE}, POPE~\citep{Li2023EvaluatingOH}, and AMBER~\citep{Wang2023AnLM}.
CHAIR and SHR evaluate hallucinations in open-ended generative tasks, POPE focuses on discriminative object-presence questions, and AMBER covers both generative and discriminative settings.
Detailed descriptions of these benchmarks and their evaluation metrics are provided in Appendix~\ref{A: exp setting}.

\textbf{Baselines.}
Following prior work~\citep{Zhao2025TellMW}, we compare \ourmethod~with three representative VAA baselines: PAI~\citep{Liu2024PayingMA}, VAF~\citep{Yin2025ClearSightVS}, and VGA~\citep{Zhao2025TellMW}.
Detailed baseline configurations are provided in Appendix~\ref{A: baseline}.
TARAC~\citep{Xie2025TARACMH} is also relevant, but we do not include it because its official implementation is not publicly available, avoiding potential discrepancies from unofficial reproduction.

\textbf{Implementation details.}
For each LVLM, we use a model-specific set of hyperparameters and keep it fixed across all benchmarks.
For LLaVA-NeXT-7B, we set the balance coefficient $\alpha$ to 0.5, the risk scale $\gamma$ to 0.5, the maximum VAA factor $m_{\text{vis}}^{\max}$ to 1.1, and the text suppression factor $m_{\text{txt}}^{\max}$ to 1.7.
For Qwen3-VL-8B, we set $\alpha$ to 0.6, $\gamma$ to 0.6, $m_{\text{vis}}^{\max}$ to 1.3, and $m_{\text{txt}}^{\max}$ to 1.3.
For InternVL3.5-8B, we set $\alpha$ to 0.8, $\gamma$ to 0.7, $m_{\text{vis}}^{\max}$ to 1.3, and $m_{\text{txt}}^{\max}$ to 1.6.
Sensitivity analysis for these hyperparameters is provided in Section~\ref{sec: sensitivity analysis}.
Following prior work~\citep{Zhao2025TellMW}, we set the layer range to $[0,16)$ for LLaVA-NeXT-7B and $[4,16)$ for Qwen3-VL-8B and InternVL3.5-8B.
For each LVLM, we tune hyperparameters on 500 validation samples from the CHAIR benchmark and use the resulting model-specific hyperparameter set unchanged across all benchmarks.

\subsection{Main Result}
\label{sec: main result}
\textbf{CHAIR evaluation.}
Table~\ref{tab:chair} reports the results on the CHAIR benchmark.
Across all three LVLMs, \ourmethod~achieves the best \emph{CHAIRs} and \emph{CHAIRi} scores, outperforming VAA baselines by a clear margin.
In particular, \ourmethod~reduces \emph{CHAIRs} by at least 3.60, 7.60, and 7.00 points on LLaVA-NeXT-7B, Qwen3-VL-8B, and InternVL3.5-8B, respectively, while also lowering instance-level hallucinations measured by \emph{CHAIRi}.
These gains are obtained without sacrificing generation quality, as the \emph{F1} scores remain comparable to or slightly better than the vanilla models.

\textbf{SHR evaluation.}
Table~\ref{tab:shr_results} reports the results on the SHR benchmark, where GPT-5-mini is used as the judge; the judge prompt is provided in Figure~\ref{fig: gpt prompt}.
Across all three LVLMs, \ourmethod~achieves the best scores on all four SHR metrics, including \emph{HSR}, \emph{HWR}, \emph{HSPI}, and \emph{HWPI}, showing consistent improvements over VAA baselines.
These results indicate that \ourmethod~effectively reduces both sentence-level and word-level hallucinations, with particularly strong gains on InternVL3.5-8B.

\begin{table*}[t]
\centering
\caption{Results on CHAIR benchmark. We report the best result in \textbf{bold}.}
\label{tab:chair}
\resizebox{0.85\linewidth}{!}{%
\begin{tabular}{l|ccc|ccc|ccc}
\toprule
\midrule
\multirow{2}{*}{Method} & \multicolumn{3}{c|}{LLaVA-NeXT-7B} & \multicolumn{3}{c|}{Qwen3-VL-8B} & \multicolumn{3}{c}{InternVL3.5-8B} \\
 & CHAIRs $\downarrow$ & CHAIRi $\downarrow$ & F1 $\uparrow$ & CHAIRs $\downarrow$ & CHAIRi $\downarrow$ & F1 $\uparrow$ & CHAIRs $\downarrow$ & CHAIRi $\downarrow$ & F1 $\uparrow$ \\
\midrule
Vanilla           & 33.80 & 8.46 & 71.44 & 58.80 & 10.57 & \textbf{75.29} & 41.40 & 10.80 & 74.71 \\
PAI               & 39.60 & 10.06 & \textbf{72.15} & 53.60 & 10.47 & 74.95 & 45.60 & 11.81 & 74.98 \\
VAF               & 36.40 & 9.14 & 71.66 & 53.60 & 9.75 & 75.15 & 42.20 & 11.00 & 74.58 \\
VGA               & 32.40 & 9.90 & 71.26 & 58.40 & 10.50 & 74.73 & 44.20 & 11.57 & 74.68 \\
\rowcolor{lg}
Ours  & \textbf{28.80} & \textbf{7.66} & 71.18 & \textbf{46.00} & \textbf{8.38} & 75.22 & \textbf{34.40} & \textbf{9.34} & \textbf{75.32}\\
\midrule
\bottomrule
\end{tabular}
}
\end{table*}
\begin{table*}[t]
\centering
\caption{Results on SHR benchmark. GPT-5-mini is the judge. We report the best result in \textbf{bold}.}
\label{tab:shr_results}
\resizebox{0.95\linewidth}{!}{%
\begin{tabular}{l|cccc|cccc|cccc}
\toprule
\midrule
\multirow{2}{*}{Method} & \multicolumn{4}{c|}{LLaVA-NeXT-7B} & \multicolumn{4}{c|}{Qwen3-VL-8B} & \multicolumn{4}{c}{InternVL3.5-8B} \\
 & HSR $\downarrow$ & HWR $\downarrow$ & HSPI $\downarrow$ & HWPI $\downarrow$ & HSR $\downarrow$ & HWR $\downarrow$ & HSPI $\downarrow$ & HWPI $\downarrow$ & HSR $\downarrow$ & HWR $\downarrow$ & HSPI $\downarrow$ & HWPI $\downarrow$ \\
\midrule
Vanilla & 31.40 & 32.80 & 2.71 & 48.19 & 23.10 & 24.70 & 3.94 & 72.62 & 26.10 & 29.70 & 2.15 & 37.73 \\
PAI     & 32.80 & 34.40 & 2.79 & 49.96 & 26.20 & 28.50 & 3.87 & 72.21 & 24.30 & 26.90 & 1.85 & 33.40 \\
VAF     & 31.90 & 33.50 & 2.77 & 49.30 & 24.00 & 25.70 & 4.09 & 75.78 & 25.90 & 29.20 & 2.13 & 37.01 \\
VGA     & 31.10 & 32.70 & \textbf{2.69} & 47.93 & 22.70 & 24.30 & 3.88 & 71.64 & 25.40 & 28.70 & 2.08 & 36.22 \\
\rowcolor{lg}
Ours  & \textbf{30.70} & \textbf{32.60} & \textbf{2.69} & \textbf{45.14} & \textbf{22.10} & \textbf{23.30} & \textbf{3.51} & \textbf{69.84} & \textbf{22.40} & \textbf{25.70} & \textbf{1.63} & \textbf{26.74} \\ 
\midrule
\bottomrule
\end{tabular}%
}
\vspace{-10pt}
\end{table*}
\begin{table*}[t]
\centering
\caption{Results on AMBER benchmark. We report the best result in \textbf{bold}.}
\label{tab:amber_results}
\resizebox{0.95\linewidth}{!}{%
\begin{tabular}{l|cccc|cccc|cccc}
\toprule
\midrule
\multirow{2}{*}{Method} & \multicolumn{4}{c|}{LLaVA-NeXT-7B} & \multicolumn{4}{c|}{Qwen3-VL-8B} & \multicolumn{4}{c}{InternVL3.5-8B} \\
 & CHAIR $\downarrow$ & Cover $\uparrow$ & Hal $\downarrow$ & Cog $\downarrow$ & CHAIR $\downarrow$ & Cover $\uparrow$ & Hal $\downarrow$ & Cog $\downarrow$ & CHAIR $\downarrow$ & Cover $\uparrow$ & Hal $\downarrow$ & Cog $\downarrow$ \\
\midrule
Vanilla & 7.83 & 63.87 & 49.20 & 4.33 & 7.66 & 73.59 & 59.24 & 3.75 & 7.49 & 74.24 & 62.20 & 5.95 \\
PAI & 8.29 & \textbf{64.74} & 50.90 & 4.43 & 8.23 & \textbf{74.11} & 63.39 & 3.50 & 8.51 & \textbf{74.88} & 66.00 & 5.29 \\
VAF & 7.64 & 63.24 & 47.41 & 3.70 & 7.41 & 72.75 & 56.64 & 3.38 & 7.30 & 71.64 & 53.20 & 3.99 \\
VGA & 7.65 & 62.39 & 40.74 & 3.59 & 7.97 & 73.32 & 60.31 & 3.34 & 7.36 & 73.99 & 62.09 & 5.49 \\
\rowcolor{lg}
Ours & \textbf{6.97} & 61.21 & \textbf{39.85} & \textbf{3.41} & \textbf{6.74} & 72.12 & \textbf{51.30} & \textbf{3.00} & \textbf{6.50} & 72.30 & \textbf{50.24} & \textbf{2.78}\\
\midrule
\bottomrule
\end{tabular}%
}
\vspace{-10pt}
\end{table*}

\textbf{AMBER evaluation.}
Table~\ref{tab:amber_results} reports the results on the AMBER benchmark.
AMBER jointly evaluates hallucination severity and content coverage, providing a balanced assessment of generation quality.
Across all three LVLMs, \ourmethod~achieves the lowest hallucination rate (i.e., \emph{Hal}) while incurring only marginal reductions in \emph{Cover}, yielding a favorable trade-off between hallucination mitigation and semantic coverage.
\ourmethod~also consistently improves other generative metrics, including \emph{CHAIR} and \emph{Cog}, across all models.
Full AMBER results, including discriminative tasks, are provided in Appendix~\ref{A: amber full}.
We observe smaller gains on discriminative tasks than on generative ones because discriminative settings typically require predicting a single closed-form token (e.g., a binary answer), making them closer to one-shot visual classification and leaving limited room for step-wise adaptive VAA.

\textbf{POPE evaluation.}
POPE results are summarized in Appendix~\ref{A: pope}, with performance averaged over three VQA datasets: MSCOCO, A-OKVQA, and GQA.
Similar to AMBER discriminative tasks, \ourmethod~brings modest gains on POPE because it requires predicting a single closed-form answer, leaving limited room for step-wise adaptive VAA.
Importantly, across all three LVLMs and datasets, \ourmethod~preserves the original model utility, achieving accuracy and \emph{F1} scores comparable to the vanilla baselines.
These results show that \ourmethod~does not compromise discriminative capabilities while improving open-ended generative hallucination mitigation.

\subsection{Ablation Study}
\label{sec: ablation}
\textbf{Ablation study on method components.}
We ablate the main components of \ourmethod~in Table~\ref{tab:ablation_method} of Appendix~\ref{A: ablation}.
Both step-wise VAA factor calibration and text attention suppression consistently reduce hallucination metrics across all three LVLMs, indicating that the two components contribute complementary benefits.
In particular, VAA factor calibration yields the largest improvements on both \emph{CHAIRs} and \emph{CHAIRi}, highlighting the importance of strengthening visual grounding during autoregressive generation.

\textbf{Ablation study on text suppression scope.}
We further study the scope of attention suppression in Table~\ref{tab:ablation_suppress_scope} of Appendix~\ref{A: ablation}.
Suppressing all text tokens achieves the lowest hallucination scores, but causes a noticeable drop in \emph{F1}, suggesting overly conservative generation.
In contrast, restricting suppression to text tokens provides a better trade-off, reducing hallucinations while preserving generation quality across all three LVLMs.
These results support our design choice of using text attention suppression as a lightweight complementary mechanism.

\textbf{Ablation study on grounding-vector pooling.}
We further study different pooling strategies for constructing the grounding vector $\mathbf{G}$, including max pooling, mean pooling, and top-$5$ average pooling (see Table~\ref{tab:chair_pooling} in Appendix~\ref{A: ablation}).
The results show that \ourmethod~is robust to the pooling choice across all three LVLMs.
Although max pooling is used as the default strategy, mean and top-$5$ average pooling achieve comparable performance.
This suggests that the visual grounding signal remains effective under different aggregation strategies.

\subsection{Sensitivity Analysis}
\label{sec: sensitivity analysis}

We conduct sensitivity analysis in Appendix~\ref{A: sensitivity analysis}.
Although \ourmethod~introduces several hyperparameters, its parameterization is comparable to existing VAA baselines, which also require method- or architecture-specific configurations.
For each LVLM, we tune hyperparameters on 500 validation samples from the CHAIR benchmark and then keep the resulting model-specific hyperparameter set fixed across all benchmarks.
This makes the tuning process lightweight and avoids benchmark-specific hyperparameter search.
For the balance coefficient $\alpha$ and risk scale $\gamma$, \ourmethod~shows stable performance across a wide range of values and consistently outperforms VAA baselines under all tested settings (see Tables~\ref{tab:alpha_sensitivity_chair} and~\ref{tab:risk_scale_chair}).
For the maximum VAA factor $m_{\text{vis}}^{\max}$ and the text suppression factor $m_{\text{txt}}^{\max}$, excessively large values can degrade performance, especially on Qwen3-VL (see Tables~\ref{tab:m_visual_chair} and~\ref{tab:m_text_chair}).
This is expected, as overly aggressive visual amplification or text suppression may distort the original attention distribution and cause the model to overemphasize or underutilize certain modalities, consistent with prior observations~\citep{Liu2024PayingMA}.
Within moderate ranges, however, \ourmethod~consistently reduces hallucinations while maintaining stable \emph{F1} scores, indicating robustness under reasonable intervention strengths.

\subsection{Token-level Hallucination Analysis}

We further analyze token-level hallucinations on CHAIR examples using Qwen3-VL-8B, with GPT-5-mini as the judge (see Appendix \ref{A: hallucination type}).
As shown in Table~\ref{tab:chair_overall}, \ourmethod~generates more total tokens than the vanilla model, yet reduces hallucinated tokens from 9,439 to 8,180 and lowers the hallucination ratio from 5.64\% to 4.85\%.
This suggests that \ourmethod~does not reduce hallucinations by producing shorter or overly conservative outputs.
Table~\ref{tab:chair_categories} further shows that the largest reduction comes from object hallucinations, consistent with the role of VAA in strengthening visual grounding.

\begin{wrapfigure}{r}{0.4\linewidth}
    \vspace{-10pt}
    \begin{centering}
    \includegraphics[width=\linewidth]{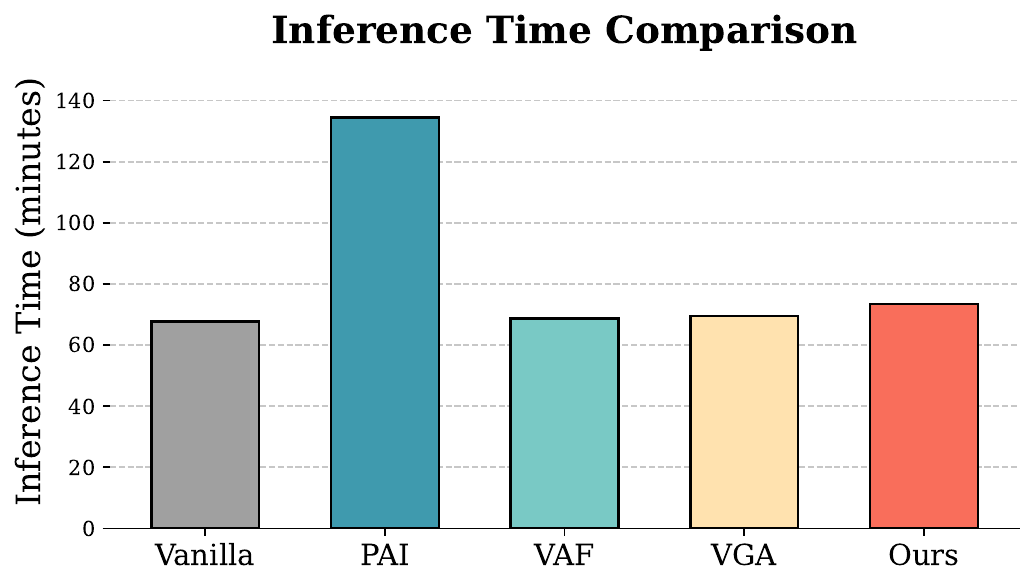}
    \caption{Inference-time comparison on the CHAIR benchmark using LLaVA-NeXT-7B on a single A100 GPU.}
    \label{fig: inference}
    \end{centering}
    \vspace{-10pt}
\end{wrapfigure}

\subsection{Inference Efficiency and Compute Resources}
\label{sec: compute resource}
As shown in Figure~\ref{fig: inference}, \ourmethod~incurs only marginal inference-time overhead compared with the vanilla model, while remaining substantially more efficient than PAI.
The overhead is lightweight: $\mathbf{G}$ is computed once during prefilling and reused across decoding steps, while each step only adds entropy computation, grounding-score lookup, VGE risk estimation, and VAA factor calibration.
\ourmethod~requires no additional training, model fine-tuning, or extra decoding passes.
All experiments on LLaVA-NeXT-7B, Qwen3-VL-8B, and InternVL3.5-8B can be run on a single NVIDIA A100 GPU with 40GB memory, making \ourmethod~practical for standard single-GPU inference-time evaluation.

\vspace{-5pt}
\section{Limitations}
\label{sec: limitation}
\textbf{Hyperparameter dependency.}
One limitation is that \ourmethod~uses model-specific hyperparameters. Future work may explore more parameter-efficient and fully adaptive calibration strategies.

\textbf{Limited gains on discriminative tasks.}
In addition, \ourmethod~is most beneficial for open-ended autoregressive generation, while gains on single-token discriminative tasks are relatively modest.
Extending SAVVA to such closed-form prediction settings remains an interesting future direction.

\vspace{-5pt}
\section{Conclusion}
We identify a dual failure pattern in prior VAA methods: a fixed amplification factor can be insufficient at some generation steps while excessive at others.
To address this issue, we propose \ourmethod, a new VAA framework that uses VGE to estimate hallucination risk and calibrate the VAA factor in a one-step-lagged manner during autoregressive generation.
By combining risk-guided, step-wise visual attention amplification with lightweight text attention suppression, \ourmethod~mitigates hallucinations while preserving generation quality.
Extensive experiments across multiple LVLMs and hallucination benchmarks demonstrate consistent improvements over existing VAA baselines.

\newpage
\section*{Impact Statement}
\label{Sec: impact statement}
SAVAA aims to improve the factual reliability of LVLMs by reducing visually unsupported content in open-ended generation.
This can benefit applications that rely on image-grounded descriptions, such as visual assistance, content understanding, and human-AI interaction.
However, hallucination mitigation should not be interpreted as a guarantee of factual correctness, and more fluent or reliable-looking outputs may increase user over-trust.
Therefore, deployment in high-stakes settings should still include task-specific safety checks, uncertainty communication, and human oversight.

\bibliography{example_paper}
\bibliographystyle{plainnat}
\newpage
\appendix

\section{Analysis of Hallucination in the Motivation Experiment}
\label{A: definition of hallucination}

\subsection{How to Measure Hallucination: GPT-5-mini as a Judge vs. Rule-based Metrics}
\begin{figure}[!htbp]
    \centering
    \includegraphics[width=\linewidth]{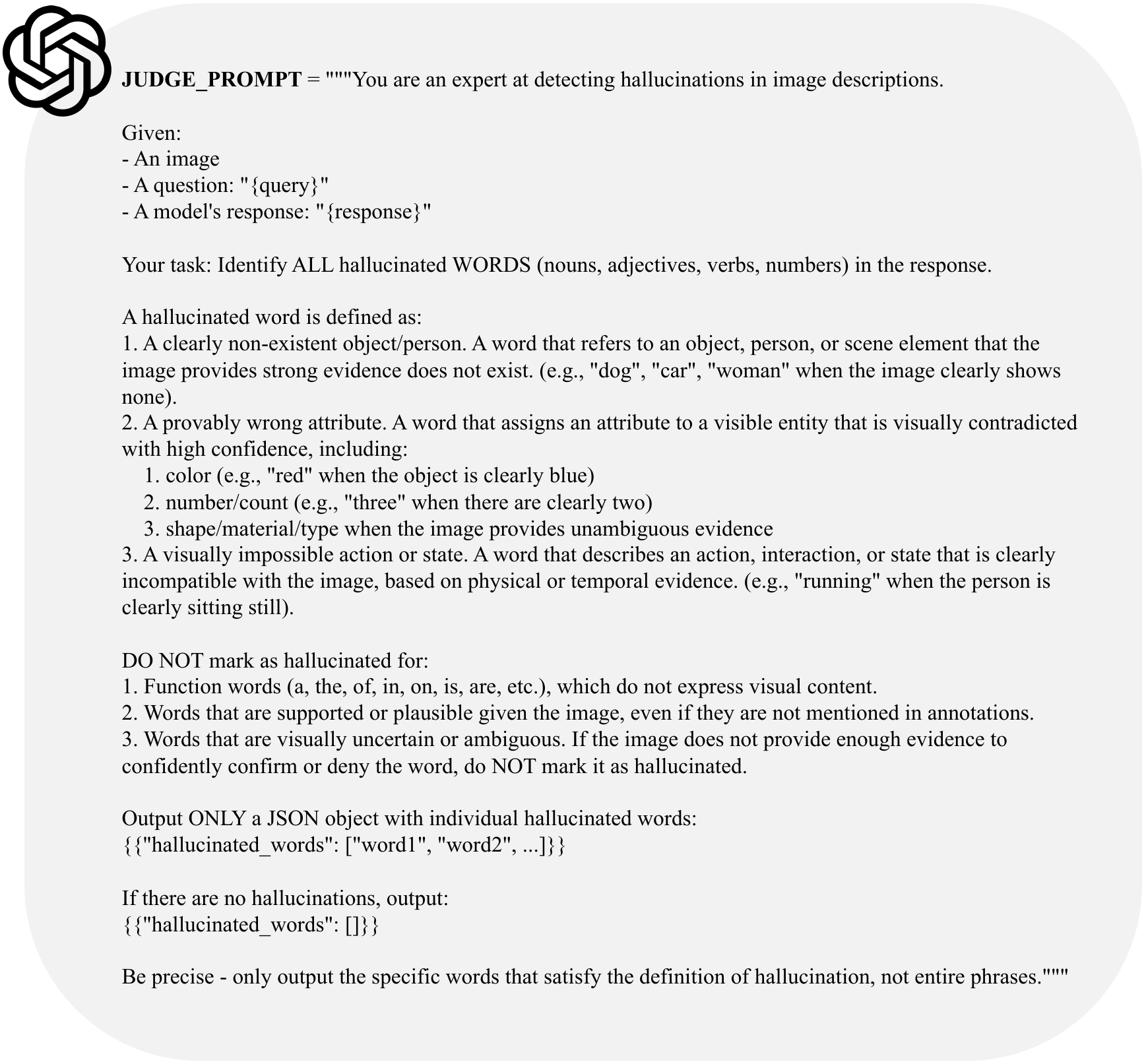}
    \caption{Prompt for GPT-5-mini to decide whether a word is hallucinated.}
    \label{fig: gpt prompt}
\end{figure}

Accurately evaluating hallucinations in LVLMs depends critically on how hallucination is defined under the target task.
Our motivation experiments and the problem setting of this paper focus on open-ended prompts such as \emph{``Describe the image''}, where models are encouraged to generate natural, narrative-style descriptions rather than strictly enumerate annotated visual entities.
Therefore, we mainly use GPT-5-mini as a token-level judge to determine whether a generated token is hallucinated; the judge prompt is provided in Figure~\ref{fig: gpt prompt}.

Prior hallucination benchmarks such as AMBER often rely on rule-based metrics that flag tokens not explicitly grounded in the provided annotations.
While effective for closed-set or object-centric evaluation, these metrics can over-penalize benign narrative elaborations in open-ended image description tasks, leading to inflated false positives.
In contrast, GPT-5-mini enables a more contextual judgment: a token is considered hallucinated only when it is inconsistent with the visual input, rather than merely absent from the annotation set.
Therefore, we use standard benchmark metrics for main evaluation and GPT-5-mini for token-level analyses in our motivation and diagnostic experiments.

\begin{figure}[!htbp]
    \begin{centering}
    \includegraphics[width=\linewidth]{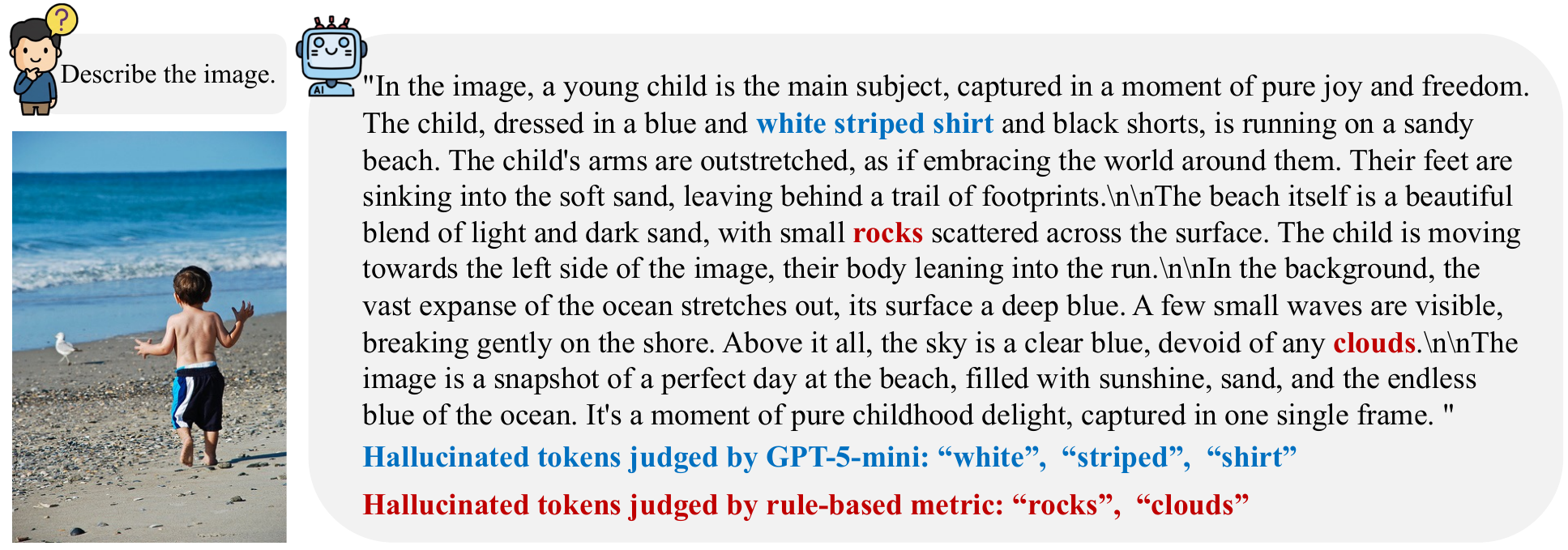}
    \caption{An example of GPT-5-mini as a judge vs. rule-based metrics for hallucination detection.}
    \label{fig: rule-based-1}
    \end{centering}
    \vspace{-10pt}
\end{figure}

\begin{figure}[!htbp]
    \begin{centering}
    \includegraphics[width=\linewidth]{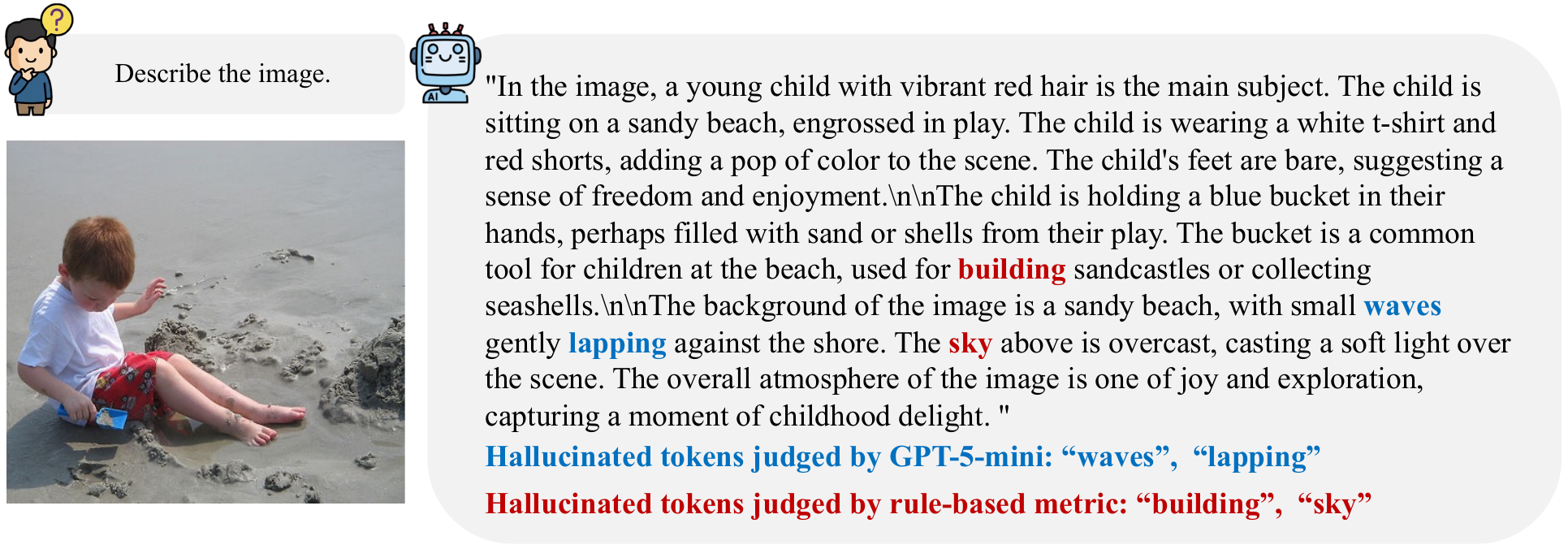}
    \caption{An example of GPT-5-mini as a judge vs. rule-based metrics for hallucination detection.}
    \label{fig: rule-based-2}
    \end{centering}
    \vspace{-10pt}
\end{figure}

Figures~\ref{fig: rule-based-1} and~\ref{fig: rule-based-2} illustrate this limitation with representative examples.
In the first example, the rule-based metric flags tokens such as \emph{rocks} and \emph{clouds} as hallucinations simply because they are absent from the annotations.
However, the generated description refers to \emph{small rocks} scattered on the beach and a sky \emph{devoid of any clouds}, both of which are visually consistent with the image.
This shows that annotation matching alone cannot assess contextual correctness in open-ended descriptions.
By contrast, GPT-5-mini does not mark these benign tokens as hallucinations and instead identifies a more salient error: the model describes the child as wearing a \emph{blue and white striped shirt}, whereas the child’s upper body is bare.

In the second example, the rule-based metric penalizes \emph{building} by incorrectly treating it as a hallucinated noun.
However, in the original phrase \emph{``used for building sandcastles''}, \emph{building} functions as a verb describing the typical use of the bucket, rather than introducing a new visual entity.
This illustrates another limitation of rule-based evaluation: surface-level token matching cannot reliably disambiguate linguistic roles or semantic functions.
As a result, harmless commonsense explanations may be mistakenly classified as hallucinations.
By contrast, GPT-5-mini recognizes the verb usage of \emph{building} and does not penalize this explanatory narration.

Importantly, GPT-5-mini can better distinguish visually inconsistent hallucinations from benign narrative elaborations, aligning more closely with human judgments in open-ended image descriptions.
Therefore, we use GPT-5-mini as the hallucination judge in our motivation experiments and diagnostic analyses, while relying on standard benchmark metrics for the main evaluation.
We acknowledge that LLM-based judges are not perfect, as LLMs may also make mistakes, and reliable hallucination measurement remains an open challenge.
Nevertheless, given the limitations of annotation-dependent rule-based metrics in open-ended generation, strong contemporary models such as GPT-5-mini provide a more semantically meaningful alternative for fine-grained token-level analysis.

\subsection{Definition of Under-amplified and Over-amplified Hallucinated Tokens}

\begin{figure}[!htbp]
    \centering
    \includegraphics[width=\linewidth]{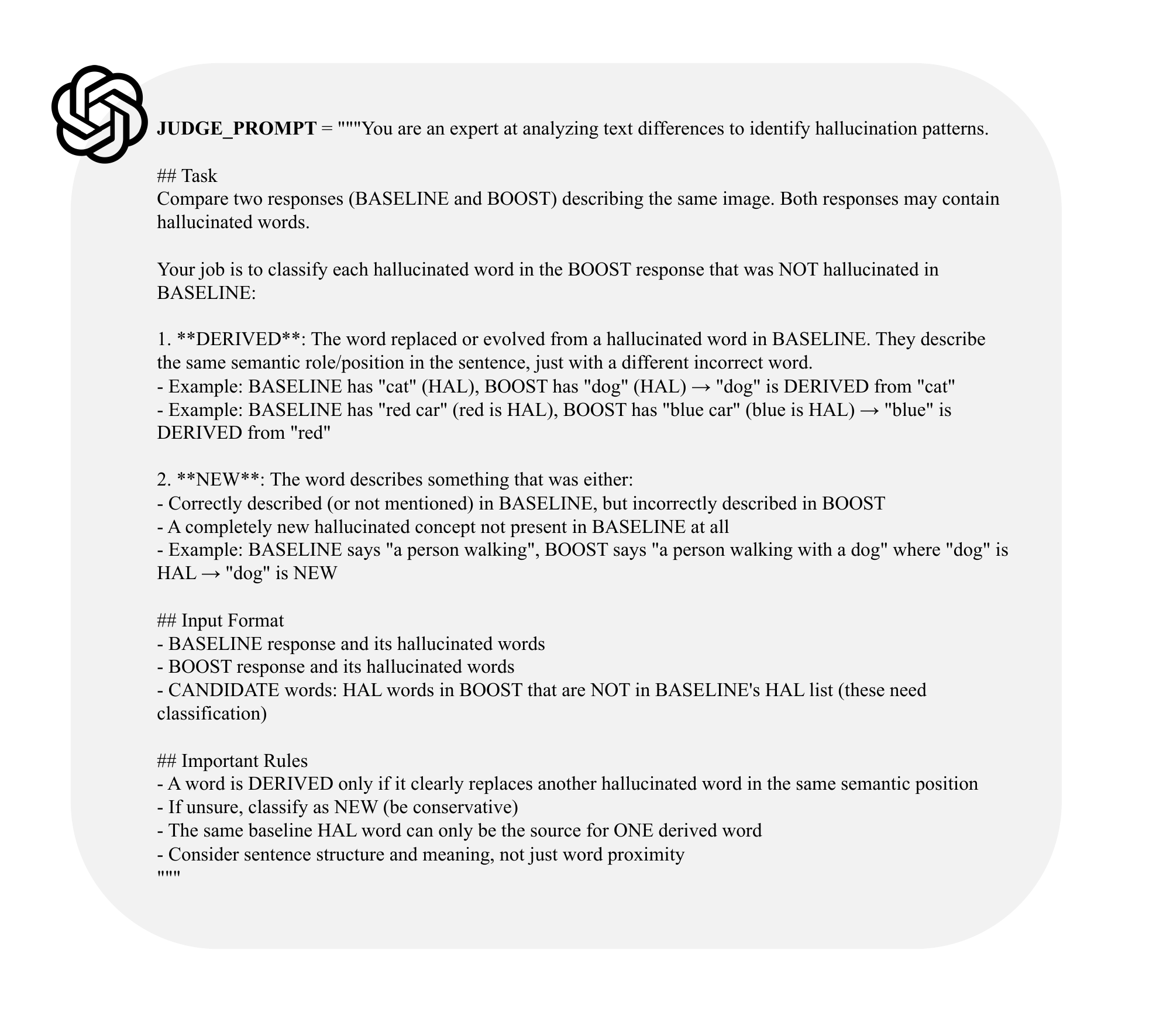}
    \caption{Prompt for GPT-5-mini to decide whether a hallucinated word is over-amplified (i.e., over-boosted) or derived from a previous hallucinated token.}
    \label{fig: gpt prompt 2}
\end{figure}

In the motivation analysis (see Figure~\ref{fig: motivation}), we characterize how uniform visual amplification affects hallucinations at the token level.
For consistency with the judge prompt in Figure~\ref{fig: gpt prompt 2}, we use \texttt{BASELINE}, \texttt{BOOST}, and \texttt{BOOST}$^{+}$ to denote the vanilla response, the response generated with a fixed VAA factor, and the response generated with a stronger fixed VAA factor, respectively.
Our analysis distinguishes (i) hallucinations newly \emph{introduced} by visual amplification, (ii) hallucinations that \emph{remain} after visual amplification, and (iii) among the remaining ones, hallucinations that can be \emph{resolved} by further increasing the VAA factor.

\textbf{Token-level hallucination sets.}
Let $\mathcal{H}_{\text{base}}$ and $\mathcal{H}_{\text{amp}}$ denote the sets of hallucinated tokens identified in the \texttt{BASELINE} and \texttt{BOOST} responses, respectively.
We first form the candidate set of newly appearing hallucinated tokens:
\begin{equation}
\mathcal{C} = \mathcal{H}_{\text{amp}} \setminus \mathcal{H}_{\text{base}}. \nonumber
\end{equation}

\textbf{Derived vs.\ truly new hallucinations.}
A key ambiguity is that a token in $\mathcal{C}$ may not represent a truly new hallucination.
It can instead be a \emph{derived} hallucination that replaces an already hallucinated token in the baseline response at the same semantic position, such as changing one incorrect attribute into another.
To disambiguate this, we use GPT-5-mini to classify each token in $\mathcal{C}$ as either \textbf{DERIVED} or \textbf{NEW}, following the prompt in Figure~\ref{fig: gpt prompt 2}.
We denote the derived subset as $\mathcal{D} \subseteq \mathcal{C}$ and the truly new subset as $\mathcal{N} = \mathcal{C} \setminus \mathcal{D}$.

\textbf{Over-amplified hallucinated tokens.}
We define \emph{over-amplified hallucinated tokens} as hallucinations that are introduced by visual amplification and are not derived from any baseline hallucination:
\begin{equation}
\mathcal{H}_{\text{over}} \triangleq \mathcal{N}. \nonumber
\end{equation}
Intuitively, $\mathcal{H}_{\text{over}}$ captures genuinely new unsupported content that emerges after applying a fixed VAA factor, rather than a surface-level mutation of an existing hallucination.

\textbf{Remaining hallucinations.}
We group together (i) hallucinations that persist from the baseline response and (ii) hallucinations that are derived mutations of baseline hallucinations, and refer to them as \emph{remaining hallucinations}:
\begin{equation}
\mathcal{H}_{\text{rem}} \triangleq 
\big(\mathcal{H}_{\text{amp}} \cap \mathcal{H}_{\text{base}}\big)\ \cup\ \mathcal{D}. \nonumber
\end{equation}
This set represents hallucinated semantics that were already present in the baseline response and are not eliminated by the current VAA factor, possibly with altered surface forms.

\textbf{Under-amplified hallucinated tokens.}
We define \emph{under-amplified hallucinated tokens} as the subset of remaining hallucinations that can be resolved by increasing the VAA factor.
Concretely, consider a stronger visual-amplification setting, denoted \texttt{BOOST}$^{+}$, with a larger fixed VAA factor.
In the motivation experiment, we set the maximum amplification factor to 1.6.
Let $\mathcal{H}_{\text{amp}^{+}}$ be the hallucinated-token set under \texttt{BOOST}$^{+}$.
A remaining hallucinated token is considered \emph{resolved} if it becomes non-hallucinatory under the stronger VAA factor, i.e., it no longer belongs to the hallucinated-token set.
We then define:
\begin{equation}
\mathcal{H}_{\text{under}} \triangleq \{\, h \in \mathcal{H}_{\text{rem}} \mid h \notin \mathcal{H}_{\text{amp}^{+}} \,\}. \nonumber
\end{equation}
Intuitively, $\mathcal{H}_{\text{under}}$ corresponds to hallucinations that the current VAA factor fails to correct, but which are fixable by further strengthening visual amplification.

\newpage
\subsection{Prompt Templates for GPT-5-mini}
\label{A: prompt template}

\newpage

\newpage
\section{Algorithm}
\label{A: algo}
\begin{algorithm}[!htbp]
\small
\caption{\ourmethod}
\label{alg:adavboost}
\begin{algorithmic}[1]
\REQUIRE Image $\mathbf{I}$, text prompt $\mathbf{x}$, LVLM $\mathcal{M}$
\REQUIRE balance coefficient $\alpha$, risk scale $\gamma$, maximum VAA factor $m^{\max}_{\text{vis}}$, maximum text suppression factor $m^{\max}_{\text{txt}}$, layer range $[L_s, L_e)$
\ENSURE Generated response $\mathbf{y} = (y_1, y_2, \ldots, y_T)$
\STATE \textit{// Prefilling stage}
\STATE $\mathbf{X}_i \gets \text{VisualEncoder}(\mathbf{I})$; $\mathbf{H}_0 \gets \mathcal{M}.\text{Prefill}(\mathbf{X}_i, \mathbf{x})$; $\mathbf{h}_i \gets \text{LMHead}(\mathbf{H}_0^{(i)}),~\forall i \in \mathcal{I}_{\mathrm{visual}}$
\STATE $\mathbf{G}[v] \gets \max_{i \in \mathcal{I}_{\mathrm{visual}}} \text{Softmax}(\mathbf{h}_i)[v],~\forall v \in \mathcal{V}$ by Eq.~\eqref{eq: grounding_vector}
\STATE \textit{// Step-wise Adaptive VAA factor calibration}
\STATE $r_0 \gets 0$ \COMMENT{Initialize risk}
\FOR{$t = 1, \ldots, T$}
    \STATE $m_t \gets 1 + (m^{\max}_{\text{vis}} - 1) \cdot r_{t-1}$ by Eq.~\eqref{eq: vaa_factor}
    \FOR{$\ell = 1, \ldots, L$}
        \STATE $\mathbf{Z}_\ell^{(t)} \gets \text{PreSoftmaxAttentionScores}(\ell)$
        \IF{$L_s \le \ell < L_e$}
            \STATE $\mathbf{Z}_\ell^{(t)}[-1, \mathcal{I}_{\mathrm{visual}}] \gets \mathbf{Z}_\ell^{(t)}[-1, \mathcal{I}_{\mathrm{visual}}] \cdot m_t$ by Eq.~\eqref{eq: visual_amplification}
            \STATE $\mathbf{Z}_\ell^{(t)}[-1, \mathcal{I}_{\mathrm{text}}] \gets \mathbf{Z}_\ell^{(t)}[-1, \mathcal{I}_{\mathrm{text}}] / m^{\max}_{\text{txt}}$ by Eq.~\eqref{eq: text_suppression}
        \ENDIF
    \ENDFOR
    \STATE $\mathbf{z}_t \gets \text{LMHead}(\mathbf{h}_t^{(L)})$;  $\mathbf{p}_t \gets \text{Softmax}(\mathbf{z}_t)$; $y_t \sim \mathbf{p}_t$; Append $y_t$ to $\mathbf{y}$
    \STATE \textit{// Update hallucination risk}
    \STATE $\bar{H}_t \gets - \sum_{v \in \mathcal{V}} p_t(v)\log p_t(v) \,/\, \log |\mathcal{V}|$ by Eq.~\eqref{eq: normalized_entropy}
    \STATE $v_t^* \gets \arg\max_{v \in \mathcal{V}} \mathbf{z}_t[v]$; $G_t \gets \mathbf{G}[v_t^*]$ by Eq.~\eqref{eq: step-wise-G}
    \STATE $\text{VGE}_t \gets \alpha \cdot \bar{H}_t + (1-\alpha)\cdot (1 - G_t)$ by Eq.~\eqref{eq: vge}
    \STATE $r_t \gets \min(1, \text{VGE}_t / \gamma)$ by Eq.~\eqref{eq: risk_score}
\ENDFOR
\STATE \textbf{Return} $\mathbf{y}$
\end{algorithmic}
\end{algorithm}

\section{Detailed Experiment Setting}
\subsection{Benchmarks and Evaluation Metrics}
\label{A: exp setting}
\textbf{CHAIR}.
CHAIR \citep{Rohrbach2018ObjectHI} is a widely used benchmark for assessing object-level hallucinations in image captioning.
It relies on ground-truth annotations of 80 object categories provided by the MSCOCO dataset \citep{Lin2014MicrosoftCC} and measures hallucination from two complementary perspectives:
$
\textit{CHAIRi}
=
\frac{\left| \mathcal{O}_{\text{hall}} \right|}
{\left| \mathcal{O}_{\text{all}} \right|},
$ and $
\textit{CHAIRs}
=
\frac{\left| \mathcal{C}_{\text{hall}} \right|}
{\left| \mathcal{C}_{\text{all}} \right|}.
$
Here, $\mathcal{O}_{\text{hall}}$ denotes the set of hallucinated object mentions
that appear in the generated captions but are not present in the ground-truth
annotations, and $\mathcal{O}_{\text{all}}$ denotes the set of all object
mentions in the generated captions.
$\mathcal{C}_{\text{hall}}$ denotes the set of generated captions that contain at
least one hallucinated object, while $\mathcal{C}_{\text{all}}$ denotes the set
of all generated captions.
Lower values of both \emph{CHAIRi} and \emph{CHAIRs} indicate fewer object
hallucinations and better factual grounding in image descriptions.
Following standard practice for evaluating LVLMs on CHAIR \citep{Liu2024PayingMA, Xie2025TARACMH, Zhao2025TellMW}, we randomly sample 500 images from the COCO2014 validation set and prompt the models with “Please describe the image in detail”. 
Following \citet{Zhao2025TellMW}, we additionally select another 500 samples for hyperparameter tuning.

\textbf{SHR}.
SHR \citep{Zhao2023BeyondHE} is designed to evaluate hallucinations in
fine-grained image descriptions.
It is built on a subset of 200 images from the VG-100K dataset \citep{Krishna2016VisualGC}, each of which is associated with detailed annotations including object categories, attributes, spatial relationships, and bounding boxes. 
To evaluate LVLMs, models are instructed to generate detailed image captions using the prompt \emph{``Please describe this image in detail''}.
The generated captions are subsequently assessed by a large language model
(we use GPT-5-mini as a judge), which compares each sentence against the corresponding visual annotations and categorizes it as \emph{correct}, \emph{hallucination}, or \emph{cannot judge} when the description is subjective or ambiguous.
Based on these sentence-level judgments, SHR reports a set of fine-grained
metrics and we use four representative ones, including \emph{hallucinated sentences per image} (HSPI), \emph{hallucinated words per image} (HWPI), \emph{hallucination sentence ratio} (HSR), and \emph{hallucination word ratio} (HWR).
Lower values of these metrics indicate better factual alignment between generated captions and visual content.

\textbf{POPE}. POPE \citep{Li2023EvaluatingOH} provides a discriminative evaluation protocol for
object hallucinations in multimodal large language models.
Instead of relying on free-form image captions, POPE reformulates hallucination
assessment as a binary visual question answering task, where models are asked
to answer simple yes-or-no questions about the existence of specific objects in
an image (e.g., \emph{``Is there a chair in the image?''}).
This design enables a more controlled evaluation of object-level hallucinations.
Model performance is reported using standard classification metrics, including
Accuracy, Precision, Recall, and F1 score, offering a comprehensive view of both hallucination reduction and overall discriminative capability.
Following \citet{Yin2025ClearSightVS, Zhao2025TellMW}, we report the averaged results across MSCOCO \citep{Lin2014MicrosoftCC}, A-OKVQA \citep{Schwenk2022AOKVQAAB} and GQA \citep{Hudson2019GQAAN} on the POPE benchmark.

\textbf{AMBER}. AMBER \citep{Wang2023AnLM} is a benchmark designed to assess
hallucinations in LVLMs from both generative and discriminative perspectives.
Compared to CHAIR, the AMBER dataset \citep{Wang2023AnLM} covers a broader range of visual contexts,
features more balanced object categories, and includes richer object-level
annotations within each image.
AMBER reports multiple evaluation metrics to characterize different aspects of
hallucination behavior.
Specifically, the \emph{CHAIR} score in AMBER shares the same definition as
\emph{CHAIRi}, while \emph{Cover} measures object coverage and is analogous to recall.
The \emph{Hal} metric quantifies the proportion of generated responses that
contain hallucinated content, and \emph{Cog} reflects the degree to which model
outputs are driven by language priors or commonsense reasoning rather than
grounded visual evidence.
Following the default AMBER evaluation protocol, we prompt the model with
\emph{``Describe this image.''}.

\subsection{Baselines}
\label{A: baseline}
For all the baseline methods we compare, we use default parameters they provided in the paper.

\textbf{PAI.}
The VAA coefficient is set to $\alpha=0.2$.
The classifier-free guidance $\gamma$ is set to 1.1 for LLaVA-NeXT-7B, and 1.05 for Qwen3-VL-8B and InternVL3.5-8B, considering their higher visual token ratios.
VAA is applied to layers [0, 32) for LLaVA-NeXT-7B and [4, 36) for Qwen3-VL-8B and InternVL3.5-8B, consistent with their respective model depths.

\textbf{VAF.}
The visual enhancement parameter is set to $\texttt{enh\_para}=1.15$, and the system prompt suppression parameter is set to $\texttt{sup\_para}=0.90$.
VAF is applied to the middle fusion layers [9, 15) across all models.

\textbf{VGA.}
We set the attention coefficient to $\beta=0.2$.
We enable similarity-based head balancing and disable attention normalization, following the recommended configuration.
The top-$k$ parameter for entropy-based salience estimation is set to $k=10$.
Attention modulation is applied to layers [0, 15) for LLaVA-NeXT-7B and [4, 15) for Qwen3-VL-8B and InternVL3.5-8B.

\newpage
\section{Additional Experiments}
\subsection{Full Experiment Results on AMBER}
\label{A: amber full}
\begin{table*}[!htbp]
\centering
\caption{Results on AMBER benchmark. The AMBER metric is calculated as $(1 - \text{CHAIR} + \text{F1})/2$.}
\label{tab:amber_full}
\resizebox{\linewidth}{!}{%
\begin{tabular}{ll|cccc|ccccc}
\toprule
\midrule
MLLM & Method & CHAIR $\downarrow$ & Cover $\uparrow$ & Hal $\downarrow$ & Cog $\downarrow$ & Acc. $\uparrow$ & Prec. $\uparrow$ & Rec. $\uparrow$ & F1 $\uparrow$ & AMBER $\uparrow$ \\
\midrule
\multirow{6}{*}{LLaVA-NeXT-7B} 
 & Vanilla & 7.83 & 63.87 & 49.20 & 4.33 & \textbf{85.28} & 90.42 & 87.03 & \textbf{88.69} & 90.43 \\
 & PAI & 8.29 & \textbf{64.74} & 50.90 & 4.43 & 83.92 & \textbf{93.61} & 81.30 & 87.02 & 89.37 \\
 & VAF & 7.64 & 63.24 & 47.41 & 3.70 & 84.24 & 90.64 & 85.00 & 87.72 & 90.04 \\
 & VGA  & 7.65 & 62.39 & 40.74 & 3.59 & 84.24 & 88.07 & \textbf{88.17} & 88.11 & 90.23 \\
 & \cellcolor{lg}{Ours} & \cellcolor{lg}{\textbf{6.97}} & \cellcolor{lg}{61.21} & \cellcolor{lg}{\textbf{39.85}} & \cellcolor{lg}{\textbf{3.41}} & \cellcolor{lg}{85.08} & \cellcolor{lg}{90.20} & \cellcolor{lg}{86.92} & \cellcolor{lg}{88.52} & \cellcolor{lg}{\textbf{90.78}} \\
\midrule
\multirow{6}{*}{Qwen3-VL-8B} 
 & Vanilla & 7.66 & 73.59 & 59.24 & 3.75 & \textbf{89.09} & \textbf{91.96} & \textbf{91.53} & \textbf{91.74} & 92.04 \\
 & PAI & 8.23 & \textbf{74.11} & 63.39 & 3.50 & 88.93 & 91.72 & 91.56 & 91.63 & 91.70 \\
 & VAF & 7.41 & 72.75 & 56.64 & 3.38 & 88.07 & 91.49 & 90.38 & 90.93 & 91.76 \\
 & VGA & 7.97 & 73.32 & 60.31 & 3.34 & 88.60 & 92.17 & 90.47 & 91.31 & 91.67 \\
 & \cellcolor{lg}{Ours} & \cellcolor{lg}{\textbf{6.74}} & \cellcolor{lg}{72.12} & \cellcolor{lg}{\textbf{51.30}} & \cellcolor{lg}{\textbf{3.00}} & \cellcolor{lg}{\textbf{89.09}} & \cellcolor{lg}{\textbf{91.96}} & \cellcolor{lg}{\textbf{91.53}} & \cellcolor{lg}{\textbf{91.74}} & \cellcolor{lg}{\textbf{92.50}} \\
\midrule
\multirow{5}{*}{InternVL3.5-8B} 
 & Vanilla & 7.49 & 74.24 & 62.20 & 5.95 & 87.77 & 93.48 & 87.64 & 90.46 & 91.48 \\
 & PAI & 8.51 & \textbf{74.88} & 66.00 & 5.29 & 87.68 & 93.82 & 87.13 & 90.35 & 90.92 \\
 & VAF & 7.30 & 71.64 & 53.20 & 3.99 & 87.17 & 93.89 & 86.23 & 89.89 & 91.30 \\
 & VGA & 7.36 & 73.99 & 62.09 & 5.49 & \textbf{88.06} & \textbf{93.99} & \textbf{87.57} & 90.66 & 91.65 \\
 & \cellcolor{lg}{Ours} & \cellcolor{lg}{\textbf{6.50}} & \cellcolor{lg}{72.30} & \cellcolor{lg}{\textbf{50.24}} & \cellcolor{lg}{\textbf{2.78}} & \cellcolor{lg}{87.77} & \cellcolor{lg}{93.48} & \cellcolor{lg}{87.64} & \cellcolor{lg}{90.46} & \cellcolor{lg}{\textbf{91.98}} \\
\midrule
\bottomrule
\end{tabular}
}
\end{table*}

\subsection{Full Experiment Results on POPE}
\label{A: pope}
\begin{table*}[!htbp]
  \centering
  \caption{Results on POPE. The results reported as the average performance across the MSCOCO, A-OKVQA, and GQA datasets.}
  \label{tab:pope}
  \resizebox{\linewidth}{!}{%
  \begin{tabular}{l|cccc|cccc|cccc}
  \toprule
  \multirow{2}{*}{Method} & \multicolumn{4}{c|}{LLaVA-NeXT-7B} & \multicolumn{4}{c|}{Qwen3-VL-8B} & \multicolumn{4}{c}{InternVL3.5-8B} \\
  & Acc.$\uparrow$ & Prec.$\uparrow$ & Rec.$\uparrow$ & F1$\uparrow$ & Acc.$\uparrow$ & Prec.$\uparrow$ & Rec.$\uparrow$ & F1$\uparrow$ & Acc.$\uparrow$ & Prec.$\uparrow$ & Rec.$\uparrow$ & F1$\uparrow$ \\
  \midrule
  \multicolumn{13}{c}{Random} \\
  \midrule
  Vanilla         & \textbf{91.34} & \textbf{89.71} & 93.80 & \textbf{91.57} & \textbf{93.24} & 95.04 & 91.37 & 93.09 & 90.96 & 89.66 & 92.64 & 91.08 \\
  PAI              & 90.36 & 87.27 & 95.13 & 90.85 & 93.23 & 93.89 & 92.62 & \textbf{93.17} & 90.17 & 87.66 & 93.62 & 90.51 \\
  VAF              & 90.24 & 86.46 & \textbf{95.89} & 90.80 & 93.10 & 93.14 & \textbf{93.22} & 93.09 & 91.70 & 89.08 & \textbf{95.17} & 91.99 \\
  VGA              & 90.83 & 89.31 & 93.29 & 91.08 & 92.82 & 94.30 & 91.29 & 92.68 & \textbf{92.32} & \textbf{90.58} & 94.53 & \textbf{92.48} \\
  \rowcolor{lg} Ours             & 91.33 & 89.70 & 93.78 & 91.55 & 93.16 & \textbf{95.18} & 91.06 & 92.99 & 91.26 & 88.52 & 94.97 & 91.59 \\
  \midrule
  \multicolumn{13}{c}{Popular} \\
  \midrule
  Vanilla         & \textbf{86.01} & 81.83 & 93.82 & \textbf{87.15} & \textbf{88.48} & \textbf{86.61} & 91.35 & 88.80 & 82.85 & \textbf{77.60} & 92.59 & 84.35 \\
  PAI              & 84.09 & 78.83 & 95.13 & 85.88 & 88.45 & 85.78 & 92.60 & \textbf{88.92} & 81.28 & 75.18 & 93.53 & 83.31 \\
  VAF              & 84.64 & 78.86 & \textbf{95.89} & 86.33 & 87.14 & 83.45 & \textbf{93.11} & 87.88 & 82.84 & 76.45 & \textbf{95.17} & 84.74 \\
  VGA              & 83.45 & 78.94 & 93.22 & 85.12 & 87.46 & 84.97 & 91.29 & 87.90 & \textbf{83.31} & 77.31 & 94.50 & \textbf{85.00} \\
  \rowcolor{lg} Ours             & \textbf{86.01} & \textbf{81.85} & 93.80 & 87.14 & 88.35 & \textbf{86.61} & 91.04 & 88.64 & 82.64 & 76.32 & 94.91 & 84.55 \\
  \midrule
  \multicolumn{13}{c}{Adversarial} \\
  \midrule
  Vanilla         & \textbf{80.18} & \textbf{74.47} & 93.82 & \textbf{82.72} & \textbf{84.11} & 80.25 & 91.35 & \textbf{85.24} & 78.68 & \textbf{72.71} & 92.59 & 81.30 \\
  PAI              & 77.23 & 70.95 & 95.13 & 80.93 & 83.55 & 78.96 & 92.60 & 85.00 & 77.22 & 70.71 & 93.53 & 80.44 \\
  VAF              & 77.81 & 71.01 & \textbf{95.89} & 81.36 & 82.39 & 77.17 & \textbf{93.11} & 84.17 & 78.48 & 71.55 & \textbf{95.17} & 81.61 \\
  VGA              & 77.46 & 71.87 & 93.22 & 80.77 & 83.05 & 78.85 & 91.29 & 84.40 & \textbf{78.71} & 72.04 & 94.50 & \textbf{81.66} \\
  \rowcolor{lg} Ours             & 80.15 & 74.46 & 93.80 & 82.70 & 84.09 & \textbf{80.38} & 91.04 & 85.17 & 78.46 & 71.61 & 94.91 & 81.54 \\
  \bottomrule
  \end{tabular}%
  }
  \end{table*}

\newpage
\subsection{Ablation Study}
\label{A: ablation}

\begin{table*}[!htbp]
\centering
\caption{Ablation study on method components on the CHAIR benchmark.}
\label{tab:ablation_method}
\resizebox{\linewidth}{!}{%
\begin{tabular}{l|ccc|ccc|ccc}
\toprule
\midrule
\multirow{2}{*}{Method} & \multicolumn{3}{c|}{LLaVA-NeXT-7B} & \multicolumn{3}{c|}{Qwen3-VL-8B} & \multicolumn{3}{c}{InternVL3.5-8B} \\
 & CHAIRs $\downarrow$ & CHAIRi $\downarrow$ & F1 $\uparrow$ & CHAIRs $\downarrow$ & CHAIRi $\downarrow$ & F1 $\uparrow$ & CHAIRs $\downarrow$ & CHAIRi $\downarrow$ & F1 $\uparrow$ \\
\midrule
Vanilla & 33.80 & 8.46 & \textbf{71.44} & 58.80 & 10.57 & \textbf{75.29} & 41.40 & 10.80 & 74.71 \\
+ Text Suppression  & 31.60 & 8.00 & 71.24 & 52.40 & 9.98 & 74.39 & 39.20 & 10.45 & 74.70 \\
+ Visual Amplification & \textbf{28.80} & \textbf{7.66} & 71.18 & \textbf{46.00} & \textbf{8.38} & 75.22 & \textbf{34.40} & \textbf{9.34} & \textbf{75.32}\\
\midrule
\bottomrule
\end{tabular}
}
\end{table*}

\begin{table*}[!htbp]
\centering
\caption{Ablation study on text suppression scope on the CHAIR benchmark.}
\label{tab:ablation_suppress_scope}
\resizebox{\linewidth}{!}{%
\begin{tabular}{l|ccc|ccc|ccc}
\toprule
\midrule
\multirow{2}{*}{Suppression Scope} & \multicolumn{3}{c|}{LLaVA-NeXT-7B} & \multicolumn{3}{c|}{Qwen3-VL-8B} & \multicolumn{3}{c}{InternVL3.5-8B} \\
 & CHAIRs$\downarrow$ & CHAIRi$\downarrow$ & F1$\uparrow$ & CHAIRs$\downarrow$ & CHAIRi$\downarrow$ & F1$\uparrow$ & CHAIRs$\downarrow$ & CHAIRi$\downarrow$ & F1$\uparrow$ \\
\midrule
All Text Tokens & \textbf{26.40} & 7.89 & 69.63 & \textbf{41.80} & \textbf{7.79} & 74.18 & 38.40 & 10.49 & 75.08 \\
Text Output Only & 33.20 & 9.05 & \textbf{71.49} & 55.20 & 10.97 & 74.46 & 38.80 & 10.75 & 74.44 \\
System Prompt Only & 31.80 & 9.60 & 70.69 & 50.00 & 8.93 & 74.63 & \textbf{33.60} & 9.95 & 74.74 \\
Text Input Only & 28.80 & \textbf{7.66} & 71.18 & 46.00 & 8.38 & \textbf{75.22} & 34.40 & \textbf{9.34} & \textbf{75.32} \\
\midrule
\bottomrule
\end{tabular}%
}
\end{table*}

\begin{table*}[!htbp]
\centering
\caption{Ablation study on pooling strategies for the grounding vector on the CHAIR benchmark.}
\label{tab:chair_pooling}
\resizebox{\linewidth}{!}{%
\begin{tabular}{l|ccc|ccc|ccc}
\toprule
\midrule
\multirow{2}{*}{Pooling} 
& \multicolumn{3}{c|}{LLaVA-NeXT-7B} 
& \multicolumn{3}{c|}{Qwen3-VL-8B} 
& \multicolumn{3}{c}{InternVL3.5-8B} \\
& CHAIRs $\downarrow$ & CHAIRi $\downarrow$ & F1 $\uparrow$ 
& CHAIRs $\downarrow$ & CHAIRi $\downarrow$ & F1 $\uparrow$ 
& CHAIRs $\downarrow$ & CHAIRi $\downarrow$ & F1 $\uparrow$ \\
\midrule
Max
& \textbf{28.80} & 7.66 & \textbf{71.18} 
& 46.00 & 8.38 & 75.22 
& \textbf{34.40} & \textbf{9.34} & \textbf{75.32} \\

Mean 
& 29.40 & 7.54 & \textbf{71.18} 
& \textbf{43.20} & \textbf{8.10} & \textbf{76.41} 
& 34.60 & 10.06 & 74.23 \\

Top-5 avg 
& 29.00 & \textbf{7.40} & 71.17 
& 45.80 & 8.61 & 75.94 
& 35.80 & 9.98 & 74.70 \\
\midrule
\bottomrule
\end{tabular}
}
\end{table*}

\subsection{Sensitivity Analysis}
\label{A: sensitivity analysis}

\subsubsection{Balance Coefficient}
\begin{table}[!htbp]
\centering
\caption{Sensitivity analysis on balance coefficient $\alpha$ on the CHAIR benchmark.}
\label{tab:alpha_sensitivity_chair}
\resizebox{\linewidth}{!}{%
\begin{tabular}{c|ccc|ccc|ccc}
\toprule
\midrule
\multirow{2}{*}{$\alpha$} & \multicolumn{3}{c|}{LLaVA-NeXT-7B} & \multicolumn{3}{c|}{Qwen3-VL-8B} & \multicolumn{3}{c}{InternVL3.5-8B} \\
 & CHAIRs$\downarrow$ & CHAIRi$\downarrow$ & F1$\uparrow$ & CHAIRs$\downarrow$ & CHAIRi$\downarrow$ & F1$\uparrow$ & CHAIRs$\downarrow$ & CHAIRi$\downarrow$ & F1$\uparrow$ \\
\midrule
0.5 & 28.80 & 7.66 & 71.18 & \textbf{46.00} & 8.58 & 75.44 & 35.80 & 9.66 & 74.31 \\
0.6 & 29.00 & \textbf{7.52} & 71.06 & \textbf{46.00} & \textbf{8.38} & 75.22 & 35.40 & 9.87 & 74.94 \\
0.7 & \textbf{28.60} & 7.59 & 70.95 & 48.80 & 8.80 & 75.38 & 34.80 & 9.46 & 75.12 \\
0.8 & 29.40 & 7.96 & 70.85 & 49.40 & 9.25 & \textbf{75.92} & 34.40 & 9.34 & \textbf{75.32} \\
0.9 & 30.40 & 7.98 & 71.18 & 46.60 & 9.39 & 75.69 & 35.60 & 9.78 & 75.02 \\
1.0 & 29.80 & 8.03 & \textbf{71.69} & 50.20 & 9.31 & 74.66 & \textbf{34.20} & \textbf{9.26} & 75.08 \\
\midrule
\bottomrule
\end{tabular}%
}
\end{table}

\subsubsection{Risk Scale}
\begin{table}[!htbp]
\centering
\caption{Sensitivity analysis on risk scale $\gamma$ on the CHAIR benchmark.}
\label{tab:risk_scale_chair}
\resizebox{\linewidth}{!}{%
\begin{tabular}{c|ccc|ccc|ccc}
\toprule
\midrule
\multirow{2}{*}{$\gamma$} & \multicolumn{3}{c|}{LLaVA-NeXT-7B} & \multicolumn{3}{c|}{Qwen3-VL-8B} & \multicolumn{3}{c}{InternVL3.5-8B} \\
 & CHAIRs$\downarrow$ & CHAIRi$\downarrow$ & F1$\uparrow$ & CHAIRs$\downarrow$ & CHAIRi$\downarrow$ & F1$\uparrow$ & CHAIRs$\downarrow$ & CHAIRi$\downarrow$ & F1$\uparrow$ \\
\midrule
0.5 & \textbf{28.80} & 7.66 & \textbf{71.18} & \textbf{43.60} & \textbf{8.19} & 75.08 & 35.20 & 9.97 & 74.68 \\
0.6 & 29.00 & 7.85 & 70.97 & 46.00 & 8.38 & 75.22 & 34.80 & 10.24 & \textbf{75.46} \\
0.7 & 29.00 & \textbf{7.53} & 71.13 & 48.20 & 8.81 & \textbf{75.50} & 34.40 & 9.34 & 75.32 \\
0.8 & 29.60 & 7.69 & 70.75 & 47.80 & 9.21 & 75.40 & \textbf{32.60} & \textbf{8.88} & 75.08 \\
0.9 & 30.60 & 8.06 & 70.36 & 51.20 & 8.57 & 75.40 & 34.00 & 9.86 & 75.03 \\
1.0 & 29.60 & 7.91 & 71.10 & 50.00 & 10.79 & 74.59 & 34.60 & 9.25 & 74.83 \\
\midrule
\bottomrule
\end{tabular}%
}
\end{table}

\newpage
\subsubsection{Maximum VAA Factor}
\begin{table}[!htbp]
\centering
\caption{Sensitivity analysis on maximum VAA factor $m_{\text{vis}}^{\max}$ on the CHAIR benchmark.}
\label{tab:m_visual_chair}
\resizebox{\linewidth}{!}{%
\begin{tabular}{c|ccc|ccc|ccc}
\toprule
\midrule
\multirow{2}{*}{$m_{\text{vis}}^{\max}$} & \multicolumn{3}{c|}{LLaVA-NeXT-7B} & \multicolumn{3}{c|}{Qwen3-VL-8B} & \multicolumn{3}{c}{InternVL3.5-8B} \\
 & CHAIRs$\downarrow$ & CHAIRi$\downarrow$ & F1$\uparrow$ & CHAIRs$\downarrow$ & CHAIRi$\downarrow$ & F1$\uparrow$ & CHAIRs$\downarrow$ & CHAIRi$\downarrow$ & F1$\uparrow$ \\
\midrule
1.1 & \textbf{28.80} & 7.66 & \textbf{71.18} & 48.80 & 8.31 & \textbf{75.40} & 35.40 & 9.40 & 75.06 \\
1.2 & 29.00 & 7.75 & 71.05 & 51.20 & 8.56 & \textbf{75.40} & 33.40 & \textbf{9.08} & 75.49 \\
1.3 & 30.60 & 7.47 & 70.66 & 46.00 & 8.38 & 75.22 & 34.40 & 9.34 & 75.32 \\
1.4 & 30.60 & \textbf{7.43} & 70.67 & 39.60 & 7.97 & 74.98 & 34.60 & 9.58 & \textbf{75.54} \\
1.5 & 31.60 & 7.61 & 70.94 & 33.60 & 7.39 & 71.92 & 35.40 & 10.22 & 75.03 \\
1.6 & 31.40 & 7.49 & 70.83 & 25.20 & 4.57 & 68.71 & \textbf{33.20} & 9.36 & 74.90 \\
1.7 & 31.40 & 8.52 & 70.22 & 21.00 & 4.49 & 65.56 & 35.20 & 9.63 & 75.32 \\
1.8 & 31.00 & 8.60 & 70.86 & 15.80 & 3.73 & 62.48 & 35.40 & 9.59 & 74.52 \\
1.9 & 30.20 & 8.00 & 70.54 & 13.80 & 4.01 & 58.88 & 33.60 & 9.74 & 75.38 \\
2.0 & 30.20 & 8.33 & 70.16 & \textbf{7.40} & \textbf{2.74} & 55.88 & 35.20 & 10.58 & 74.61 \\
\midrule
\bottomrule
\end{tabular}%
}
\end{table}

\subsubsection{Maximum Text Suppression Factor}
\begin{table}[!htbp]
\centering
\caption{Sensitivity analysis on text suppression factor $m_{\text{txt}}^{\max}$ on the CHAIR benchmark.}
\label{tab:m_text_chair}
\resizebox{\linewidth}{!}{%
\begin{tabular}{c|ccc|ccc|ccc}
\toprule
\midrule
\multirow{2}{*}{$m_{\text{txt}}^{\max}$} & \multicolumn{3}{c|}{LLaVA-NeXT-7B} & \multicolumn{3}{c|}{Qwen3-VL-8B} & \multicolumn{3}{c}{InternVL3.5-8B} \\
 & CHAIRs$\downarrow$ & CHAIRi$\downarrow$ & F1$\uparrow$ & CHAIRs$\downarrow$ & CHAIRi$\downarrow$ & F1$\uparrow$ & CHAIRs$\downarrow$ & CHAIRi$\downarrow$ & F1$\uparrow$ \\
\midrule
1.1 & 35.40 & 8.69 & \textbf{72.40} & 53.80 & 10.48 & 74.81 & 41.20 & 11.08 & 74.28 \\
1.2 & 35.60 & 9.09 & 71.08 & 53.60 & 10.15 & 75.00 & 39.20 & 10.79 & 74.65 \\
1.3 & 33.40 & 8.52 & 71.14 & 46.00 & 8.38 & \textbf{75.22} & 40.00 & 10.49 & 75.00 \\
1.4 & 33.00 & 8.92 & 71.35 & 41.20 & 8.37 & 74.36 & 38.80 & 10.14 & 75.03 \\
1.5 & 33.00 & 8.77 & 70.55 & 42.20 & 7.89 & 73.54 & 35.80 & 9.63 & 74.64 \\
1.6 & 30.00 & 8.09 & 71.53 & 41.80 & 8.55 & 72.84 & 34.40 & 9.34 & \textbf{75.32} \\
1.7 & 28.80 & 7.66 & 71.18 & 36.80 & 7.78 & 72.88 & 33.60 & 9.54 & 74.81 \\
1.8 & \textbf{28.00} & \textbf{7.15} & 71.00 & 36.80 & 8.31 & 71.41 & \textbf{29.80} & \textbf{9.06} & 74.09 \\
1.9 & 28.60 & 7.33 & 70.27 & 35.80 & 7.76 & 72.39 & 33.20 & 9.40 & 74.76 \\
2.0 & 33.20 & 9.49 & 70.53 & \textbf{35.60} & \textbf{7.05} & 73.63 & 32.60 & 9.39 & 73.59 \\
\midrule
\bottomrule
\end{tabular}%
}
\end{table}

\subsection{Types of Hallucinations Remaining after \ourmethod}
\label{A: hallucination type}
\begin{table}[!htbp]
\centering
\caption{Overall hallucination statistics on the CHAIR benchmark with 500 randomly sampled images for Qwen3-VL-8B. GPT-5-mini is used as a judge for counting hallucinated tokens.}
\label{tab:chair_overall}
\begin{tabular}{lccc}
\toprule
\midrule
Method & Total Tokens & Hallucinated Tokens & Hallucination Ratio \\
\midrule
Vanilla  & 167284 & 9439 & 5.64\% \\
\rowcolor{lg} Ours  & 168503 & 8180 & 4.85\% \\
\midrule
$\Delta$ & +1219 & -1259  & -0.79\% \\
\midrule
\bottomrule
\end{tabular}
\end{table}

\begin{table}[!htbp]
\centering
\caption{Categories of hallucinated tokens on the CHAIR benchmark with 500 randomly sampled images for Qwen3-VL-8B. Generated captions are analyzed by GPT-5-mini, which classifies each hallucinated token into five categories: Object, Attribute, Relation, Counting, and Other.}
\label{tab:chair_categories}
\begin{tabular}{lcccccc}
\toprule
\midrule
Method & Object & Attribute & Relation & Counting & Other & Total \\
\midrule
Vanilla  & 5516  & 2247 & 679 & 363 & 634 & 9439 \\
\rowcolor{lg} Ours  & 4489  & 2172 & 614 & 345 & 560 & 8180 \\
\midrule
$\Delta$ & -1027 & -75  & -65 & -18 & -74 & -1259 \\
\midrule
\bottomrule
\end{tabular}
\end{table}


\newpage

\end{document}